\documentclass[journal]{IEEEtran}

\usepackage[utf8]{inputenc}

\usepackage{times}
\usepackage{graphicx}
\usepackage{subfigure} 
\usepackage{algorithm}
\usepackage{algorithmic}
\usepackage{amsmath}
\usepackage{amsfonts}
\usepackage{amssymb}
\allowdisplaybreaks[0]
\usepackage{enumitem}
\usepackage[super]{nth}
\usepackage{microtype}
\usepackage{todonotes}
\usepackage[numbers,sort&compress]{natbib}
\usepackage{flushend}

\PassOptionsToPackage{hyphens}{url}
\usepackage[colorlinks]{hyperref}
\usepackage{color}
\definecolor{mydarkblue}{rgb}{0,0.08,0.45}
\hypersetup{ %
    pdftitle={},
    pdfauthor={},
    pdfsubject={},
    pdfkeywords={},
    pdfborder=0 0 0,
    pdfpagemode=UseNone,
    colorlinks=true,
    linkcolor=mydarkblue,
    citecolor=mydarkblue,
    filecolor=mydarkblue,
    urlcolor=mydarkblue,
    pdfview=FitH}

\usepackage{cleveref}[2012/02/15]
\crefformat{footnote}{#2\footnotemark[#1]#3}

\usepackage[english]{babel}
\addto\extrasenglish{%
}

\newcommand{\noskip}{%
  \setlength\itemsep{3pt}%
  \setlength\parsep{0pt}%
  \setlength\partopsep{0pt}%
  \setlength\parskip{0pt}%
  \setlength\topskip{0pt}%
}

\begin{document}
\title{LSTM: A Search Space Odyssey}

\author{Klaus~Greff,
        Rupesh~K.~Srivastava,
        Jan~Koutn\'{i}k,
        Bas~R.~Steunebrink,
        J\"{u}rgen~Schmidhuber%
\thanks{\copyright 2016 IEEE. Personal use of this material is permitted. Permission from IEEE must be
obtained for all other uses, in any current or future media, including reprinting/republishing this material for advertising or promotional purposes, creating new collective works, for resale or redistribution to servers or lists, or reuse of any copyrighted
component of this work in other works.
Manuscript  received  May  15,  2015;  revised  March  17,  2016;  accepted  June  9,  2016.  Date  of  publication  July 8,  2016;  date of  current  version  June  20,  2016. 
DOI: \href{https://doi.org/10.1109/TNNLS.2016.2582924}{10.1109/TNNLS.2016.2582924}

This research was supported by the Swiss National Science Foundation grants
``Theory and Practice of Reinforcement Learning 2'' (\#138219) and ``Advanced Reinforcement Learning'' (\#156682), 
and by EU projects ``NASCENCE'' (FP7-ICT-317662), ``NeuralDynamics'' (FP7-ICT-270247) and WAY (FP7-ICT-288551).}%
\thanks{K. Greff, R. K. Srivastava, J. Kout\'{i}k, B. R. Steunebrink and J. Schmidhuber are with the Istituto Dalle Molle di studi sull'Intelligenza Artificiale (IDSIA), the Scuola universitaria professionale della Svizzera italiana (SUPSI), and the Università della Svizzera italiana (USI).}%
\thanks{Author e-mails addresses: \{klaus, rupesh, hkou, bas, juergen\}@idsia.ch}%
}

\markboth{Transactions on Neural Networks and Learning Systems}%
{Greff \MakeLowercase{\textit{et al.}}: LSTM: A Search Space Odyssey}

\maketitle

\begin{abstract}
Several variants of the  Long Short-Term Memory (LSTM) architecture for recurrent neural networks have been proposed since its inception in 1995.
In recent years, these networks have become the state-of-the-art models for a variety of machine learning problems.
This has led to a renewed interest in understanding the role and utility of various computational components of typical LSTM variants. 
In this paper, we present the first large-scale analysis of eight LSTM variants on three representative tasks: speech recognition, handwriting recognition, and polyphonic music modeling.
The hyperparameters of all LSTM variants for each task were optimized separately using random search, and their importance was assessed using the powerful fANOVA framework.
In total, we summarize the results of 5400 experimental runs ($\approx 15$ years of CPU time), which makes our study the largest of its kind on LSTM networks.
Our results show that none of the variants can improve upon the standard LSTM architecture significantly, and demonstrate the forget gate and the output activation function to be its most critical components. 
We further observe that the studied hyperparameters are virtually independent and derive guidelines for their efficient adjustment.
\end{abstract} 

\begin{IEEEkeywords}
Recurrent neural networks, Long Short-Term Memory, LSTM, sequence learning, random search, fANOVA.
\end{IEEEkeywords}


\section{Introduction}
Recurrent neural networks with Long Short-Term Memory (which we will concisely refer to as LSTMs) have emerged as an effective and scalable model for several learning problems related to sequential data. 
Earlier methods for attacking these problems have either been tailored towards a specific problem or did not scale to long time dependencies. 
LSTMs on the other hand are both general and effective at capturing long-term temporal dependencies. 
They do not suffer from the optimization hurdles that plague simple recurrent networks (SRNs) \cite{Hochreiter1991, Hochreiter2001} and have been used to advance the state-of-the-art for many difficult problems. 
This includes handwriting recognition \cite{Graves2009,Pham2013,Doetsch2014} and generation \cite{Graves2013d}, language modeling \cite{Zaremba2014} and translation \cite{Luong2014}, acoustic modeling of speech \cite{Sak2014}, speech synthesis \cite{Fan2014}, protein secondary structure prediction \cite{Sonderby2014}, analysis of audio \cite{Marchi2014}, and video data \cite{Donahue2014} among others.

\begin{figure*}[t]
\centering
\includegraphics[width=0.9\textwidth]{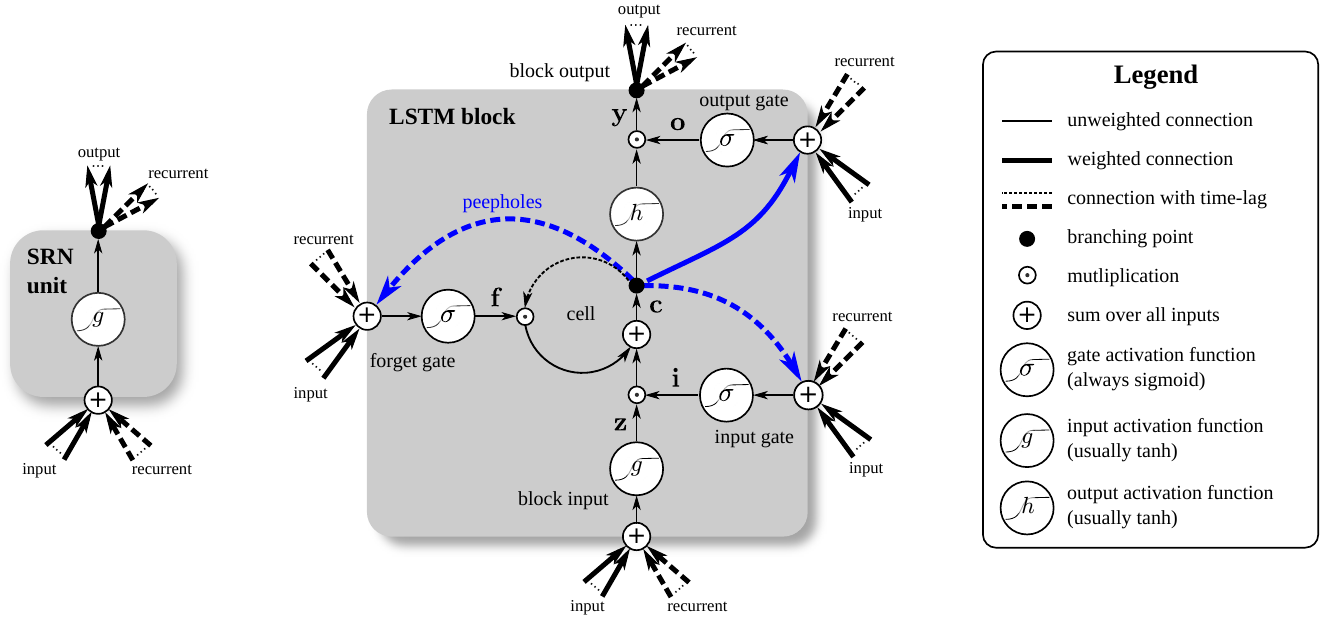}
\caption{Detailed schematic of the Simple Recurrent Network (SRN) unit (left) and a Long Short-Term Memory block (right) as used in the hidden layers of a recurrent neural network.}
\label{fig:lstm}
\end{figure*}

The central idea behind the LSTM architecture is a memory cell which can maintain its state over time, and non-linear gating units which regulate the information flow into and out of the cell.
Most modern studies incorporate many improvements that have been made to the LSTM architecture since its original formulation \cite{Hochreiter1995,Hochreiter1997}.
However, LSTMs are now applied to many learning problems which differ significantly in scale and nature from the problems that these improvements were initially tested on.
A systematic study of the utility of various computational components which comprise LSTMs (see \autoref{fig:lstm}) was missing. This paper fills that gap and systematically addresses the open question of improving the LSTM architecture.

We evaluate the most popular LSTM architecture (\emph{vanilla LSTM}; \autoref{sec:lstm}) and eight different variants thereof on three benchmark problems: acoustic modeling, handwriting recognition, and polyphonic music modeling. 
Each variant differs from the vanilla LSTM by a single change. 
This allows us to isolate the effect of each of these changes on the performance of the architecture. 
Random search \citep{Anderson1953, Solis1981, Bergstra2012} is used to find the best-performing hyperparameters for each variant on each problem, enabling a reliable comparison of the performance of different variants.
We also provide insights gained about hyperparameters and their interaction using fANOVA \citep{Hutter2014}. 


\section{Vanilla LSTM}
\label{sec:lstm}

The LSTM setup most commonly used in literature was originally described by \citet{Graves2005}. We refer to it as \emph{vanilla LSTM} and use it as a reference for comparison of all the variants.
The vanilla LSTM incorporates changes by \citet{Gers1999} and \citet{Gers2000} into the original LSTM~\cite{Hochreiter1997} and uses full gradient training. \autoref{sec:history} provides descriptions of these major LSTM changes.

A schematic of the vanilla LSTM block can be seen in \autoref{fig:lstm}.
It features three gates (input, forget, output), block input, a single cell (the Constant Error Carousel), an output activation function, and peephole connections\footnote{Some studies omit peephole connections, described in Section \ref{sec:peephole}.}. 
The output of the block is recurrently connected back to the block input and all of the gates.

\pagebreak

\subsection{Forward Pass}
\label{lstm_forward}
Let $\mathbf{x}^t$ be the input vector at time $t$, $N$ be the number of LSTM blocks and $M$ the number of inputs. Then we get the following weights for an LSTM layer:
\begin{itemize}
    \noskip
    \item Input weights: $\mathbf{W}_z$, $\mathbf{W}_i$, $\mathbf{W}_f$, $\mathbf{W}_o \in \mathbb{R}^{N \times M}$
    \item Recurrent weights: $\mathbf{R}_z$, $\mathbf{R}_i$, $\mathbf{R}_f$, $\mathbf{R}_o$ $\in \mathbb{R}^{N \times N}$
    \item Peephole weights: $\mathbf{p}_i$, $\mathbf{p}_f$, $\mathbf{p}_o$ $\in \mathbb{R}^{N}$
    \item Bias weights: $\mathbf{b}_z$, $\mathbf{b}_i$, $\mathbf{b}_f$, $\mathbf{b}_o$ $\in \mathbb{R}^{N}$
\end{itemize}

Then the vector formulas for a vanilla LSTM layer forward pass can be written as:
\begin{align*}
  \bar{\mathbf{z}}^t &= \mathbf{W}_z \mathbf{x}^t  + \mathbf{R}_z \mathbf{y}^{t-1}  + \mathbf{b}_z\\
    \mathbf{z}^t &= g(\bar{\mathbf{z}}^t) & \textit{block input}\\
    \bar{\mathbf{i}}^t &= \mathbf{W}_i \mathbf{x}^t + \mathbf{R}_i \mathbf{y}^{t-1}  + \mathbf{p}_i \odot \mathbf{c}^{t-1} + \mathbf{b}_i\\
    \mathbf{i}^t &= \sigma(\bar{\mathbf{i}}^t) & \textit{input gate}\\
    \bar{\mathbf{f}}^t &= \mathbf{W}_f \mathbf{x}^t + \mathbf{R}_f \mathbf{y}^{t-1}  + \mathbf{p}_f \odot \mathbf{c}^{t-1} + \mathbf{b}_f\\
    \mathbf{f}^t &= \sigma(\bar{\mathbf{f}}^t) & \textit{forget gate}\\
     \mathbf{c}^t &= \mathbf{z}^t \odot \mathbf{i}^t + \mathbf{c}^{t-1} \odot \mathbf{f}^t  & \textit{cell}\\
    \bar{\mathbf{o}}^t &= \mathbf{W}_o \mathbf{x}^t + \mathbf{R}_o \mathbf{y}^{t-1}  + \mathbf{p}_o \odot \mathbf{c}^{t} + \mathbf{b}_o\\
    \mathbf{o}^t &= \sigma(\bar{\mathbf{o}}^t) & \textit{output gate}\\
     \mathbf{y}^t &= h(\mathbf{c}^t) \odot \mathbf{o}^t & \textit{block output}
\end{align*}
Where $\sigma$, $g$ and $h$ are point-wise non-linear activation functions. 
The {\it logistic sigmoid} ($\sigma(x) = \frac{1}{1+e^{-x}}$) is used as gate activation function and the {\it hyperbolic tangent} ($g(x) = h(x) = \tanh(x)$) is usually used as the block input and output activation function.
Point-wise multiplication of two vectors is denoted by $\odot$.

\subsection{Backpropagation Through Time}
The deltas inside the LSTM block are then calculated as:
\begin{align*}
    \mathbf{\delta y}^t &= \Delta^t + \mathbf{R}_z^T \mathbf{\delta z}^{t+1}  +
                                      \mathbf{R}_i^T \mathbf{\delta i}^{t+1}  +
                                      \mathbf{R}_f^T \mathbf{\delta f}^{t+1}  +
                                      \mathbf{R}_o^T \mathbf{\delta o}^{t+1} \\
    \delta \bar{\mathbf{o}}^t &= \mathbf{\delta y}^t \odot h(\mathbf{c}^t) \odot \sigma'(\bar{\mathbf{o}}^t) \\
    \mathbf{\delta c}^t &= \mathbf{\delta y}^t \odot \mathbf{o}^t \odot h'(\mathbf{c}^t) +
                          \mathbf{p}_o \odot \delta \bar{\mathbf{o}}^t +
                          \mathbf{p}_i \odot \delta \bar{\mathbf{i}}^{t+1}\\
                        &\quad+\mathbf{p}_f \odot \delta \bar{\mathbf{f}}^{t+1} +
                          \mathbf{\delta c}^{t+1} \odot \mathbf{f}^{t+1} \\
    \delta \bar{\mathbf{f}}^t &= \mathbf{\delta c}^t \odot \mathbf{c}^{t-1} \odot \sigma'(\bar{\mathbf{f}}^t) \\
    \delta \bar{\mathbf{i}}^t &= \mathbf{\delta c}^t \odot \mathbf{z}^{t} \odot \sigma'(\bar{\mathbf{i}}^t) \\
    \delta \bar{\mathbf{z}}^t &= \mathbf{\delta c}^t \odot \mathbf{i}^{t} \odot g'(\bar{\mathbf{z}}^t) \\
\end{align*}

Here $\Delta^t$ is the vector of deltas passed down from the layer
above.
If $E$ is the loss function it formally corresponds to $\frac{\partial E}{\partial\mathbf{y}^{t}}$, but not including the recurrent dependencies.
The deltas for the inputs are only needed if there is a layer below that needs training, and can be computed as follows:

\[
\mathbf{\delta x}^t = \mathbf{W}_z^T \delta \bar{\mathbf{z}}^t +
                      \mathbf{W}_i^T \delta \bar{\mathbf{i}}^t +
                      \mathbf{W}_f^T \delta \bar{\mathbf{f}}^t +
                      \mathbf{W}_o^T \delta \bar{\mathbf{o}}^t
\]

Finally, the gradients for the weights are calculated as follows, where
$\mathbf{\star}$ can be any of $\{\bar{\mathbf{z}}, \bar{\mathbf{i}}, \bar{\mathbf{f}}, \bar{\mathbf{o}}\}$, and $\langle \star_1, \star_2 \rangle$ denotes the outer product of two vectors:

\begin{align*}
    \delta\mathbf{W}_\star &= \sum^T_{t=0} \langle \mathbf{\delta\star}^t, \mathbf{x}^t \rangle
    & \delta\mathbf{p}_i &= \sum^{T-1}_{t=0}  \mathbf{c}^t \odot \delta \bar{\mathbf{i}}^{t+1}
    \\
    \delta\mathbf{R}_\star &= \sum^{T-1}_{t=0} \langle \mathbf{\delta\star}^{t+1}, \mathbf{y}^t \rangle
    & \delta\mathbf{p}_f &= \sum^{T-1}_{t=0}  \mathbf{c}^t \odot \delta \bar{\mathbf{f}}^{t+1}
    \\
    \delta\mathbf{b}_\star &= \sum^{T}_{t=0} \mathbf{\delta\star}^{t}
    & \delta\mathbf{p}_o &= \sum^{T}_{t=0}    \mathbf{c}^t \odot \delta \bar{\mathbf{o}}^{t}
    \\
\end{align*}


\section{History of LSTM}
\label{sec:history}
The initial version of the LSTM block \cite{Hochreiter1995,Hochreiter1997} included (possibly multiple) cells, input and output gates, but no forget gate and no peephole connections. 
The output gate, unit biases, or input activation function were omitted for certain experiments. 
Training was done using a mixture of Real Time Recurrent Learning (RTRL) \cite{Robinson1987, Williams1989} and Backpropagation Through Time (BPTT) \cite{Werbos1988, Williams1989}. Only the gradient of the cell was propagated back through time, and the gradient for the other recurrent connections was truncated. 
Thus, that study did not use the exact gradient for training. 
Another feature of that version was the use of \emph{full gate recurrence}, which means that all the gates received recurrent inputs from all gates at the previous time-step in addition to the recurrent inputs from the block outputs.
This feature did not appear in any of the later papers. 

\subsection{Forget Gate}
The first paper to suggest a modification of the LSTM architecture introduced the forget gate \citep{Gers1999}, enabling the LSTM to reset its own state. 
This allowed learning of continual tasks such as embedded Reber grammar.

\subsection{Peephole Connections}\label{sec:peephole}
\citet{Gers2000} argued that in order to learn precise timings, the cell needs to control the gates.
So far this was only possible through an open output gate. Peephole connections (connections from the cell to the gates, blue in \autoref{fig:lstm}) were added to the architecture in order to make precise timings easier to learn.
Additionally, the output activation function was omitted, as there was no evidence that it was essential for solving the problems that LSTM had been tested on so far.

\subsection{Full Gradient}
The final modification towards the vanilla LSTM was done by \citet{Graves2005}. 
This study presented the full backpropagation through time (BPTT) training for LSTM networks with the architecture described in \autoref{sec:lstm}, and presented results on the TIMIT \cite{Garofolo1993} benchmark. Using full BPTT had the added advantage that LSTM gradients could be checked using finite differences, making practical implementations more reliable.

\subsection{Other Variants}
Since its introduction the vanilla LSTM has been the most commonly used architecture, but other variants have been suggested too.
Before the introduction of full BPTT training, \citet{Gers2002} utilized a training method based on Extended Kalman Filtering which enabled the LSTM to be trained on some pathological cases at the cost of high computational complexity.
\citet{Schmidhuber2007} proposed using a hybrid evolution-based method instead of BPTT for training but retained the vanilla LSTM architecture. 

\citet{Bayer2009} evolved different LSTM block architectures that maximize fitness on context-sensitive grammars. 
A larger study of this kind was later done by \citet{Jozefowicz2015}.
\citet{Sak2014} introduced a linear projection layer that projects the output of the LSTM layer down before recurrent and forward connections in order to reduce the amount of parameters for LSTM networks with many blocks.
By introducing a trainable scaling parameter for the slope of the gate activation functions, \citet{Doetsch2014} were able to improve the performance of LSTM on an offline handwriting recognition dataset.
In what they call \emph{Dynamic Cortex Memory}, \citet{Otte2014} improved convergence speed of LSTM by adding recurrent connections between the gates of a single block (but not between the blocks). 

\citet{Cho2014} proposed a simplified variant of the LSTM architecture called \emph{Gated Recurrent Unit} (GRU). 
They used neither peephole connections nor output activation functions, and coupled the input and the forget gate into an \emph{update gate}. 
Finally, their output gate (called \emph{reset gate}) only gates the recurrent connections to the block input ($\mathbf{W}_z$). \citet{Chung2014} performed an initial comparison between GRU and Vanilla LSTM and reported mixed results.


\section{Evaluation Setup}
The focus of our study is to empirically compare different LSTM variants, and not to achieve state-of-the-art results.
Therefore, our experiments are designed to keep the setup simple and the comparisons fair.
The vanilla LSTM is used as a baseline and evaluated together with eight of its variants.
Each variant adds, removes, or modifies the baseline in exactly one aspect, which allows to isolate their effect.
They are evaluated on three different datasets from different domains to account for cross-domain variations.

For fair comparison, the setup needs to be similar for each variant.
Different variants might require different settings of hyperparameters to give good performance, and we are interested in the best performance that can be achieved with each variant.
For this reason we chose to tune the hyperparameters like learning rate or amount of input noise individually for each variant.
Since hyperparameter space is large and impossible to traverse completely, random search was used in order to obtain good-performing hyperparameters \cite{Bergstra2012} for every combination of variant and dataset.
Random search was also chosen for the added benefit of providing enough data for analyzing the general effect of various hyperparameters on the performance of each LSTM variant (\autoref{sec:hyper-impact}).

\subsection{Datasets}
Each dataset is split into three parts: a training set, a validation set used for early stopping and for optimizing the hyperparameters, and a test set for the final evaluation.

\subsubsection*{TIMIT}
\label{sec:timit}
The TIMIT Speech corpus \cite{Garofolo1993} is large enough to be a reasonable acoustic modeling benchmark for speech recognition, yet it is small enough to keep a large study such as ours manageable.
Our experiments focus on the frame-wise classification task for this dataset, where the objective is to classify each audio-frame as one of 61 phones.\footnote{Note that in linguistics a \emph{phone} represents a distinct speech sound independent of the language. 
In contrast, a \emph{phoneme} refers to a sound that distinguishes two words \emph{in a given language} \cite{Crystal2011}.
These terms are often confused in the machine learning literature.}
From the raw audio we extract 12 Mel Frequency Cepstrum Coefficients (MFCCs) \citep{Mermelstein1976} +
energy over 25ms hamming-windows with stride of 10ms and a pre-emphasis 
coefficient of 0.97. This preprocessing is standard in speech recognition and was chosen in order to stay comparable with earlier LSTM-based results (e.g. \cite{Graves2005, Graves2008a}). The 13 coefficients along with their first and second derivatives comprise the 39 inputs to the network and were normalized to have zero mean and unit variance.

The performance is measured as classification error percentage. The training, testing, and validation sets are split in line with \citet{Halberstadt1998} into $3696$, $400$, and $192$ sequences, having $304$ frames on average.

We restrict our study to the \textit{core test set}, which is an established subset of the full TIMIT corpus, and use the splits into
training, testing, and validation sets as detailed by \citet{Halberstadt1998}. 
In short, that means we only use the core test set and drop the SA samples\footnote{The dialect sentences (the SA samples) were meant to expose the dialectal variants of the speakers and were read by all 630 speakers. We follow \cite{Halberstadt1998} and remove them because they bias the distribution of phones.} from the training set. 
The validation set is built from some of the discarded samples from the full test set. 

\subsubsection*{IAM Online}
\begin{figure}
\centering
\subfigure[]{\includegraphics[width=0.5\textwidth]{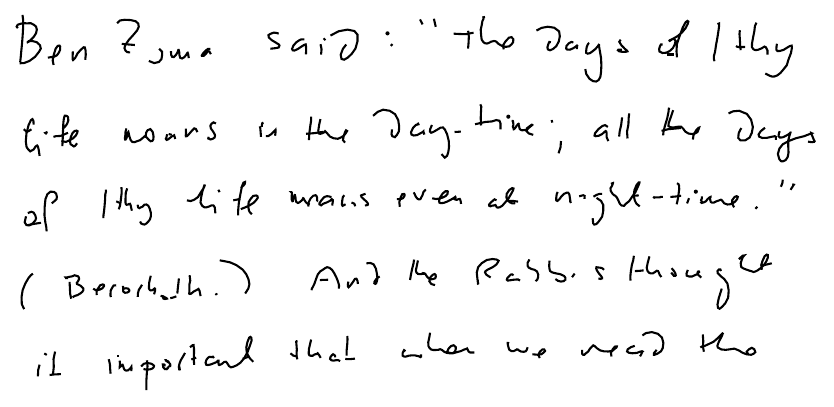}\label{fig:hwsample:strokes}}
\hfill
\subfigure[]{
\raisebox{1.1cm}{
\begin{minipage}{0.52\textwidth}
{\scriptsize\tt
Ben Zoma said: "The days of 1thy\\
life means in the day-time; all the days\\
of 1thy life means even at night-time ."\\
(Berochoth .) And the Rabbis thought\\
it important that when we read the\\
}
\end{minipage}}
\label{fig:hwsample:labels}}
\caption{(a) Example board ({\tt a08-551z}, training set) from the IAM-OnDB dataset
  and (b) its transcription into character label
  sequences.}
\label{fig:hwsample}
\end{figure}

The IAM Online Handwriting Database \cite{Liwicki2005}\footnote{The IAM-OnDB was obtained from
  \url{http://www.iam.unibe.ch/fki/databases/iam-on-line-handwriting-database}} consists of English sentences as time series of pen movements that have to be mapped to characters. 
The IAM-OnDB dataset splits into one training set, two validation sets, and one test set, having $775$,
$192$, $216$, and $544$ {\it boards} each. 
Each board, see \autoref{fig:hwsample:strokes}, contains multiple hand-written lines, which in turn consist of several strokes. We use one line per sequence, and joined the two validation sets together, so the final training, validation, and testing sets contain $5\,355$, $2\,956$ and $3\,859$ sequences respectively.

Each handwriting line is accompanied with a target character sequence, see \autoref{fig:hwsample:labels}, assembled from the following $81$~ASCII characters:
\begin{verbatim}
abcdefghijklmnopqrstuvwxyz
ABCDEFGHIJKLMNOPQRSTUVWXYZ
0123456789␣!"#&\'()*+,-./[]:;?
\end{verbatim}
The board labeled as {\tt a08-551z} (in the training set) contains a sequence of eleven percent (\%) characters that does not have an image in the strokes, and the percent character does not occur in any other board. 
That board was removed from the experiments.

We subsampled each sequence to half its length, which speeds up the training and does not harm performance. 
Each frame of the sequence is a 4-dimensional vector containing $\Delta x$, $\Delta y$ (the change in pen position),  $t$ (time since the beginning of the stroke), and a fourth dimension that contains value of one at the time of the pen lifting (a transition to the next stroke) and zeroes at all other time steps. 
Possible starts and ends of characters within each stroke are not explicitly marked. 
No additional preprocessing (like base-line straightening, cursive correction, etc.) was used.

The networks were trained using the Connectionist Temporal Classification (CTC) error function by \citet{Graves2006} with 82 outputs (81 characters plus the special empty label). We measure performance in terms of the Character Error Rate (CER) after decoding using best-path decoding \citep{Graves2006}.

\subsubsection*{JSB Chorales}
JSB Chorales is a collection of 382 four-part harmonized chorales by J. S. Bach \citep{Allan2005}, consisting of 202 chorales in major keys and 180 chorals in minor keys.
We used the preprocessed piano-rolls provided by \citet{Boulanger-Lewandowski2012}.\footnote{Available at \url{http://www-etud.iro.umontreal.ca/~boulanni/icml2012} at the time of writing.}
These piano-rolls were generated by transposing each MIDI sequence in C major or C minor and sampling frames every quarter note. The networks where trained to do next-step prediction by minimizing the negative log-likelihood. The complete dataset consists of $229$, $76$, and $77$ sequences (training, validation, and test sets respectively) with an average length of $61$.

\subsection{Network Architectures \& Training}\label{sec:arch-training}
A network with a single LSTM hidden layer and a sigmoid output layer was used for the JSB Chorales task.
Bidirectional LSTM~\cite{Graves2005} was used for TIMIT and IAM Online tasks, consisting of two hidden layers, one processing the input forwards and the other one backwards in time, both connected to a single softmax output layer.
As loss function we employed Cross-Entropy Error for TIMIT and JSB Chorales, while for the IAM Online task the Connectionist Temporal Classification (CTC) loss by \citet{Graves2006} was used.
The initial weights for all networks were drawn from a normal distribution with standard deviation of $0.1$.
Training was done using Stochastic Gradient Descent with Nesterov-style momentum \cite{Sutskever2013} with updates after each sequence. The learning rate was rescaled by a factor of $(1-\text{momentum})$. Gradients were computed using full BPTT for LSTMs \cite{Graves2005}.
Training stopped after 150 epochs or once there was no improvement on the validation set for more than fifteen epochs.

\subsection{LSTM Variants}
\label{sec:lstm-variants}
The vanilla LSTM from \autoref{sec:lstm} is referred as Vanilla~(V). 
For activation functions we follow the standard and use the logistic sigmoid for $\sigma$, and the hyperbolic tangent for both $g$ and $h$.
The derived eight variants of the V architecture are the following. We only report differences to the forward pass formulas presented in \autoref{lstm_forward}:

\begin{description}[align=right, labelwidth=1.5cm]
    \item [NIG:] No Input Gate: \( \mathbf{i}^t = \mathbf{1}\)
    \item [NFG:] No Forget Gate: \( \mathbf{f}^t = \mathbf{1}\)
    \item [NOG:] No Output Gate: \( \mathbf{o}^t = \mathbf{1}\)
    \item [NIAF:] No Input Activation Function: \(g(\mathbf{x}) = \mathbf{x}\)
    \item [NOAF:] No Output Activation Function: \(h(\mathbf{x}) = \mathbf{x}\)
    \item [CIFG:] Coupled Input and Forget Gate: 
        \( \mathbf{f}^t = \mathbf{1} - \mathbf{i}^t\)
        
    \item [NP:] No Peepholes:
\begin{align*}
        \bar{\mathbf{i}}^{t} &=\mathbf{W}_{i}\mathbf{x}^{t}+\mathbf{R}_{i}\mathbf{y}^{t-1} + \mathbf{b}_i \\
        \bar{\mathbf{f}}^{t} &=\mathbf{W}_{f}\mathbf{x}^{t}+\mathbf{R}_{f}\mathbf{y}^{t-1} + \mathbf{b}_f \\
        \bar{\mathbf{o}}^{t} &=\mathbf{W}_{o}\mathbf{x}^{t}+\mathbf{R}_{o}\mathbf{y}^{t-1} + \mathbf{b}_o
\end{align*}
        
    \item [FGR:] Full Gate Recurrence:
\begin{align*}
        \bar{\mathbf{i}}^{t} &=\mathbf{W}_{i}\mathbf{x}^{t}+\mathbf{R}_{i}\mathbf{y}^{t-1}+\mathbf{p}_{i}\odot\mathbf{c}^{t-1} + \mathbf{b}_i\\
         &\quad+ \mathbf{R}_{ii}\mathbf{i}^{t-1}+\mathbf{R}_{fi}\mathbf{f}^{t-1}+\mathbf{R}_{oi}\mathbf{o}^{t-1} \\
        \bar{\mathbf{f}}^{t} &=\mathbf{W}_{f}\mathbf{x}^{t}+\mathbf{R}_{f}\mathbf{y}^{t-1}+\mathbf{p}_{f}\odot\mathbf{c}^{t-1} + \mathbf{b}_f \\
         &\quad+ \mathbf{R}_{if}\mathbf{i}^{t-1}+\mathbf{R}_{ff}\mathbf{f}^{t-1}+\mathbf{R}_{of}\mathbf{o}^{t-1} \\
        \bar{\mathbf{o}}^{t} &=\mathbf{W}_{o}\mathbf{x}^{t}+\mathbf{R}_{o}\mathbf{y}^{t-1}+\mathbf{p}_{o}\odot\mathbf{c}^{t-1}  + \mathbf{b}_o \\
        &\quad+ \mathbf{R}_{io}\mathbf{i}^{t-1}+\mathbf{R}_{fo}\mathbf{f}^{t-1}+\mathbf{R}_{oo}\mathbf{o}^{t-1}
\end{align*}
\end{description}

The first six variants are self-explanatory. The CIFG variant uses only one gate for gating both the input and the cell recurrent self-connection -- a modification of LSTM referred to as Gated Recurrent Units (GRU) \cite{Cho2014}.
This is equivalent to setting ${\mathbf{f}_t = \mathbf{1} - \mathbf{i}_t}$ instead of learning the forget gate weights independently.
The FGR variant adds recurrent connections between all the gates as in the original formulation of the LSTM \cite{Hochreiter1997}. 
It adds nine additional recurrent weight matrices, thus significantly increasing the number of parameters.

\begin{figure*}
    \centering
    \includegraphics[width=0.95\textwidth]{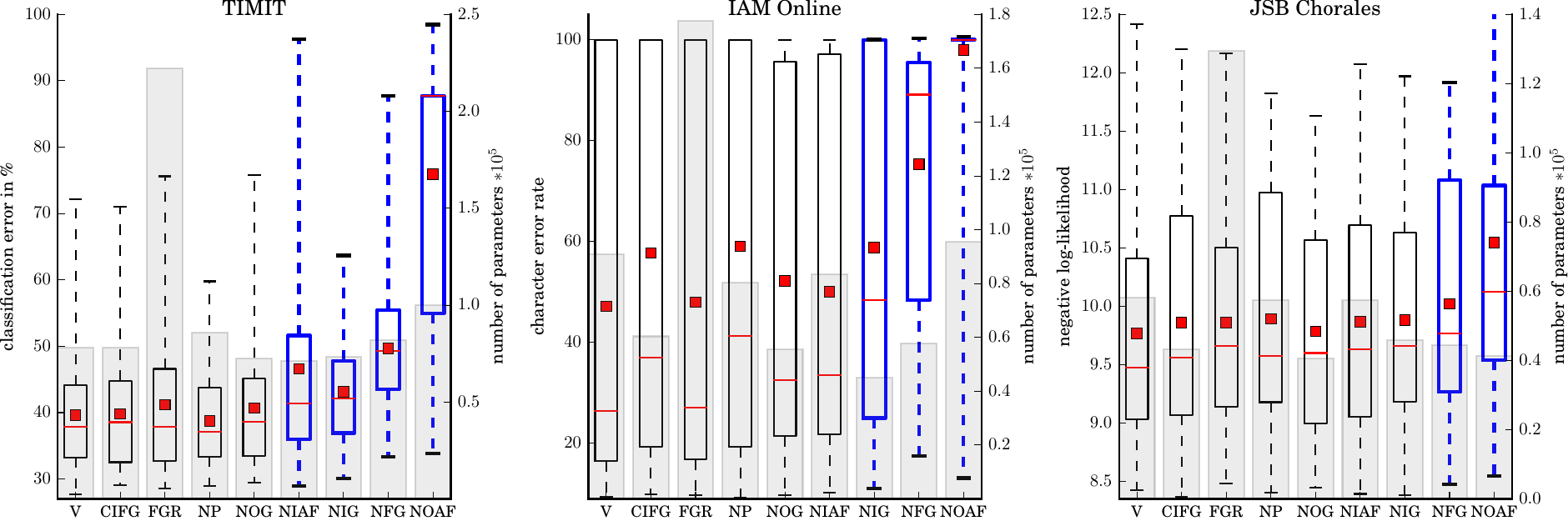}
    \includegraphics[width=0.95\textwidth]{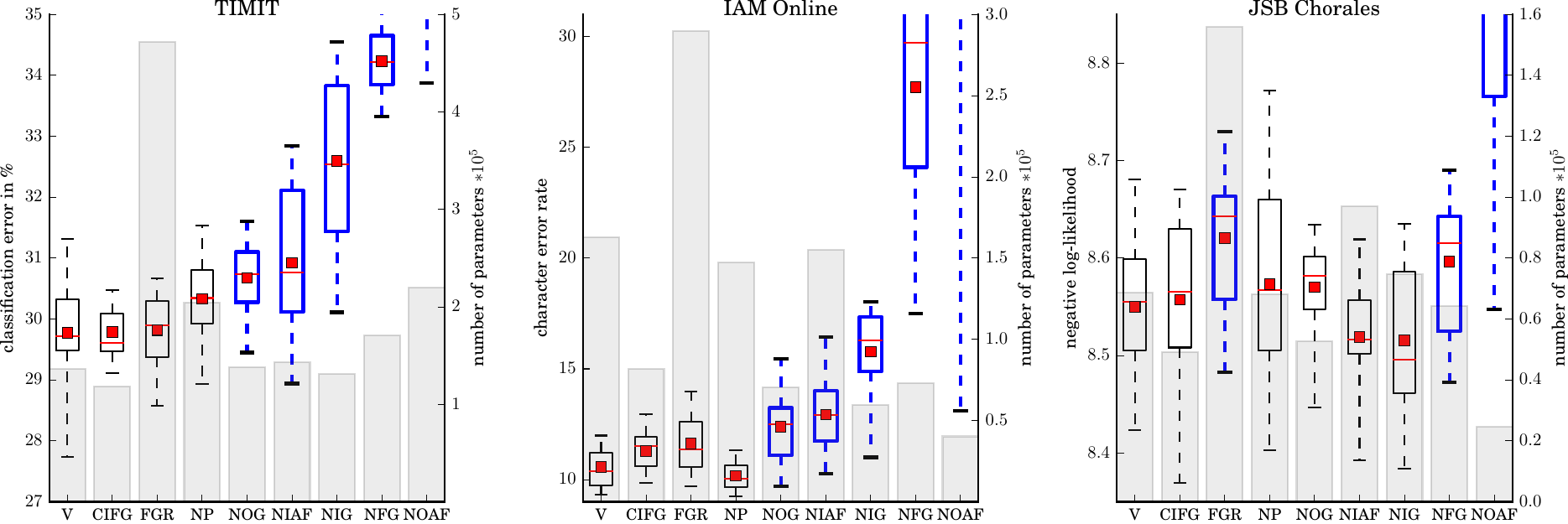}

    \caption{\emph{Test set} performance for all 200 trials (top) and for the best 10\% (bottom) trials (according to the \emph{validation set}) for each dataset and variant. Boxes show the range between the \nth{25} and the \nth{75} percentile of the data, while the whiskers indicate the whole range. The red dot represents the mean and the red line the median of the data. The boxes of variants that differ significantly from the vanilla LSTM are shown in blue with thick lines. The grey histogram in the background presents the average number of parameters for the top 10\% performers of every variant.}
    \label{fig:top20}
\end{figure*}

\subsection{Hyperparameter Search}
\label{sec:hyperparam_search}
While there are other methods to efficiently search for good hyperparameters (cf.~\cite{Snoek2012, Hutter2011}), random search has several advantages for our setting: 
it is easy to implement, trivial to parallelize, and covers the search space more uniformly, thereby improving the follow-up analysis of hyperparameter importance.

We performed $27$ random searches (one for each combination of the nine variants and three datasets).
Each random search encompasses $200$ trials for a total of $5400$ trials of randomly sampling the following hyperparameters:\\[-1em]
\begin{itemize}[leftmargin=*]
\noskip
    \item number of LSTM blocks per hidden layer: 
        log-uniform samples from $[20, 200]$;
    \item learning rate: 
        log-uniform samples from $[10^{-6}, 10^{-2}]$;
    \item momentum: 
        $1 - \text{log-uniform samples from $[0.01, 1.0]$}$;
    \item standard deviation of Gaussian input noise:
        uniform samples from $[0, 1]$.
\end{itemize}

In the case of the TIMIT dataset, two additional (boolean) hyperparameters were considered (not tuned for the other two datasets).
The first one was the choice between traditional momentum and Nesterov-style momentum \citep{Sutskever2013}. Our analysis showed that this had no measurable effect on performance so the latter was arbitrarily chosen for all further experiments. 
The second one was whether to clip the gradients to the range $[-1, 1]$. This turned out to hurt overall performance,\footnote{Although this may very well be the result of the range having been chosen too tightly.} therefore the gradients were never clipped in the case of the other two datasets.

Note that, unlike an earlier small-scale study \citep{Chung2014}, the number of parameters was not kept fixed for all variants. Since different variants can utilize their parameters differently, fixing this number can bias comparisons.


\section{Results \& Discussion}
\label{sec:results}

Each of the $5400$ experiments was run on one of 128 AMD Opteron CPUs at 2.5\,GHz and took 24.3\,h on average to complete. 
This sums up to a total single-CPU computation time of just below $15$ years.

For TIMIT the test set performance of the best trial were \textbf{29.6\%} classification error (CIFG) which is close to the best reported result of 26.9\% \cite{Graves2005}.
Our best result of \textbf{-8.38} log-likelihood (NIAF) on the JSB Chorales dataset on the other hand is well below the -5.56 from \citet{Boulanger-Lewandowski2012}. 
Best LSTM result is 26.9\% 
For the IAM Online dataset our best result was a Character Error Rate of \textbf{9.26\%} (NP) on the test set. The best previously published result is 11.5\% CER by \citet{Graves2008} using a different and much more extensive preprocessing.\footnote{Note that these numbers differ from the best test set performances that can be found in \autoref{fig:top20}. This is the case because here we only report the single best performing trial as determined on the validation set. In \autoref{fig:top20}, on the other hand, we show the test set performance of the 20 best trials for each variant.}
Note though, that the goal of this study is not to provide state-of-the-art results, but to do a fair comparison of different LSTM variants.
So these numbers are only meant as a rough orientation for the reader.

\subsection{Comparison of the Variants}
\label{sec:variant-comparison}

A summary of the random search results is shown in \autoref{fig:top20}. 
Welch's $t$-test at a significance level of $p=0.05$ was used\footnote{We applied the \emph{Bonferroni adjustment} to correct for performing eight different tests (one for each variant).} to determine whether the mean test set performance of each variant was significantly different from that of the baseline.
The box for a variant is highlighted in blue if its mean performance differs significantly from the mean performance of the vanilla LSTM.

The results in the top half of \autoref{fig:top20} represent the distribution of all 200 test set performances over the whole search space.
Any conclusions drawn from them are therefore specific to our choice of search ranges.
We have tried to chose reasonable ranges for the hyperparameters that include the best settings for each variant and are still small enough to allow for an effective search.
The means and variances tend to be rather similar for the different variants and datasets, but even here some significant differences can be found. 

In order to draw some more interesting conclusions we restrict our further analysis to the top 10\% performing trials for each combination of dataset and variant (see bottom half of \autoref{fig:top20}). 
This way our findings will be less dependent on the chosen search space and will be representative for the case of ``reasonable hyperparameter tuning efforts.''\footnote{How much effort is ``reasonable'' will still depend on the search space. If the ranges are chosen much larger, the search will take much longer to find good hyperparameters.}

The first important observation based on \autoref{fig:top20} is that removing the output activation function (NOAF) or the forget gate (NFG) significantly hurt performance on all three datasets.
Apart from the CEC, the ability to forget old information and the squashing of the cell state appear to be critical for the LSTM architecture. 
Indeed, without the output activation function, the block output can in principle grow unbounded. 
Coupling the input and the forget gate avoids this problem and might render the use of an output non-linearity less important, which could explain why GRU performs well without it.

Input and forget gate coupling (CIFG) did not significantly change mean performance on any of the datasets, although the best performance improved slightly on music modeling. Similarly, removing peephole connections (NP) also did not lead to significant changes, but the best performance improved slightly for handwriting recognition.
Both of these variants simplify LSTMs and reduce the computational complexity, so it might be worthwhile to incorporate these changes into the architecture.

Adding full gate recurrence (FGR) did not significantly change performance on TIMIT or IAM Online, but led to worse results on the JSB Chorales dataset. 
Given that this variant greatly increases the number of parameters, we generally advise against using it.
Note that this feature was present in the original proposal of LSTM \cite{Hochreiter1995, Hochreiter1997}, but has been absent in all following studies.

Removing the input gate (NIG), the output gate (NOG), and the input activation function (NIAF) led to a significant reduction in performance on speech and handwriting recognition. 
However, there was no significant effect on music modeling performance. 
A small (but statistically insignificant) average performance improvement was observed for the NIG and NIAF architectures on music modeling. 
We hypothesize that these behaviors will generalize to similar problems such as language modeling. 
For supervised learning on continuous real-valued data (such as speech and handwriting recognition), the input gate, output gate, and input activation function are all crucial for obtaining good performance.

\begin{figure*}[t]
\centering
\includegraphics[width=0.97\textwidth]{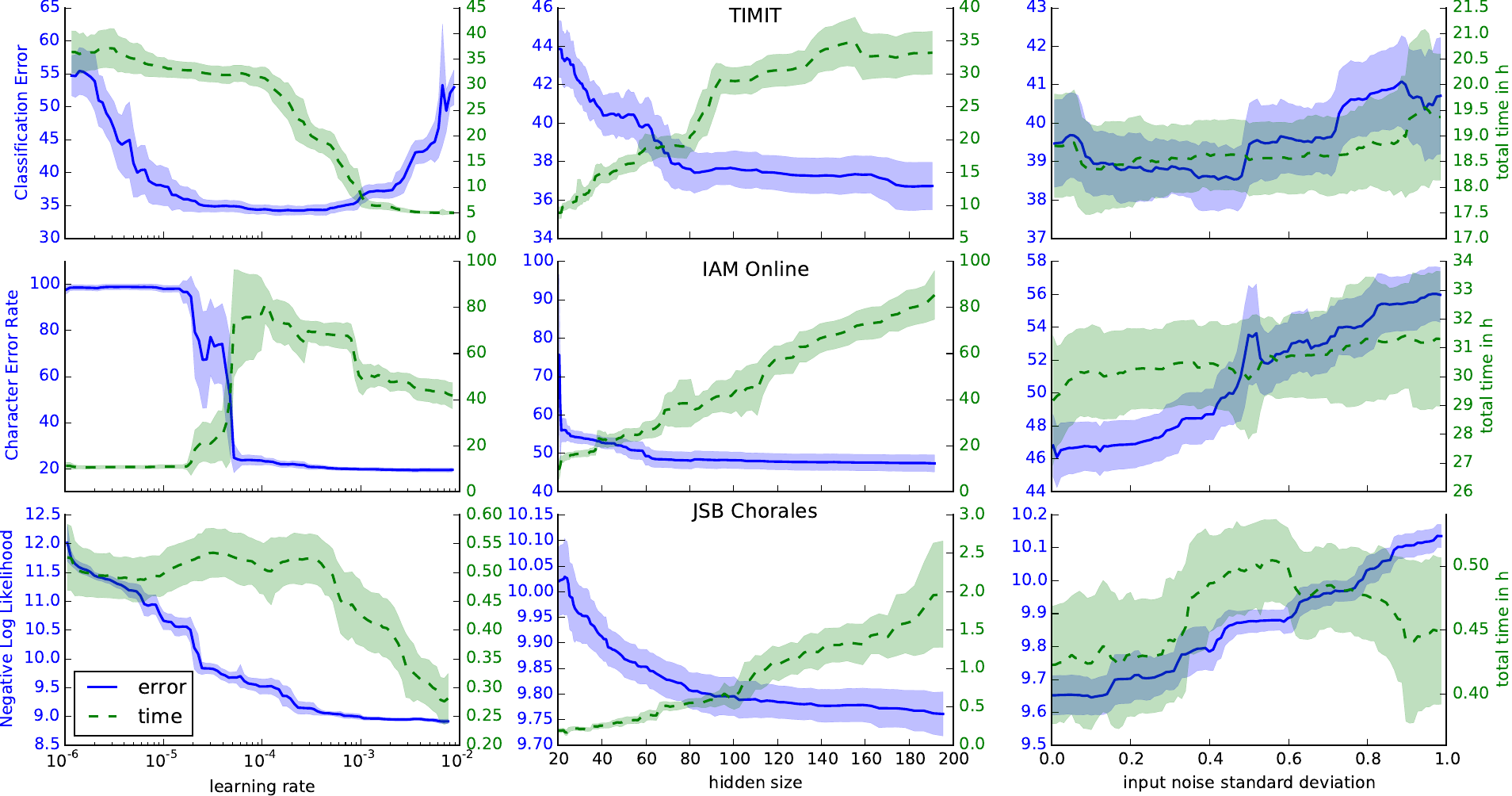}
\caption{Predicted marginal error (blue) and marginal time for different values of the \emph{learning rate}, \emph{hidden size}, and the \emph{input noise} (columns) for the test set of all three datasets (rows).
The shaded area indicates the standard deviation between the tree-predicted marginals and thus the reliability of the predicted mean performance. 
Note that each plot is for the vanilla LSTM but curves for all variants that are not significantly worse look very similar.} 
\label{fig:learning_rate}
\label{fig:hidden_size}
\label{fig:input_noise}
\end{figure*}

\subsection{Impact of Hyperparameters}
\label{sec:hyper-impact}
The fANOVA framework for assessing hyperparameter importance by \citet{Hutter2014} is based on the observation that marginalizing over dimensions can be done efficiently in regression trees. 
This allows predicting the marginal error for one hyperparameter while averaging over all the others. 
Traditionally this would require a full hyperparameter grid search, whereas here the hyperparameter space can be sampled at random. 

Average performance for any slice of the hyperparameter space is obtained by first training a regression tree and then summing over its predictions along the corresponding subset of dimensions.
To be precise, a random regression \emph{forest} of $100$ trees is trained and their prediction performance is averaged. 
This improves the generalization and allows for an estimation of uncertainty of those predictions. 
The obtained marginals can then be used to decompose the variance into additive components using the functional ANalysis Of VAriance (fANOVA) method \cite{Hooker2007} which provides an insight into the overall importance of hyperparameters and their interactions.

\subsubsection*{Learning rate}
Learning rate is the most important hyperparameter, therefore it is very important to understand how to set it correctly in order to achieve good performance. 
\autoref{fig:learning_rate} shows (in blue) how setting the learning rate value affects the predicted average performance on the test set. 
It is important to note that this is an average over all other hyperparameters and over all the trees in the regression forest. 
The shaded area around the curve indicates the standard deviation over tree predictions (not over other hyperparameters), thus quantifying the reliability of the average. 
The same is shown in green with the predicted average training time. 

The plots in \autoref{fig:learning_rate}  show that the optimal value for the learning rate is dependent on the dataset.
For each dataset, there is a large basin (up to two orders of magnitude) of good learning rates inside of which the performance does not vary much. 
A related but unsurprising observation is that there is a sweet-spot for the learning rate at the high end of the basin.\footnote{Note that it is unfortunately outside the investigated range for IAM Online and JSB Chorales. This means that ideally we should have chosen the range of learning rates to include higher values as well.}
In this region, the performance is good and the training time is small.
So while searching for a good learning rate for the LSTM, it is sufficient to do a coarse search by starting with a high value (e.g. $1.0$) and dividing it by ten until performance stops increasing.

\autoref{fig:variances} also shows that the fraction of variance caused by the learning rate is much bigger than the fraction due to interaction between learning rate and hidden layer size (some part of the ``higher order'' piece, for more see below at \textit{Interaction of Hyperparameters}). 
This suggests that the learning rate can be quickly tuned on a small network and then used to train a large one.

\subsubsection*{Hidden Layer Size}
Not surprisingly the hidden layer size is an important hyperparameter affecting the LSTM network performance.
As expected, larger networks perform better, but with diminishing returns.
It can also be seen in \autoref{fig:hidden_size} (middle, green) that the required training time increases with the network size.
Note that the scale here is \emph{wall-time} and thus factors in both the increased computation time for each epoch as well as the convergence speed.

\subsubsection*{Input Noise}
Additive Gaussian noise on the inputs, a traditional regularizer for neural networks, has been used for LSTM as well. 
However, we find that not only does it almost always hurt performance, it also slightly increases training times. 
The only exception is TIMIT, where a small dip in error for the range of $[0.2, 0.5]$ is observed. 

\subsubsection*{Momentum}
One unexpected result of this study is that momentum affects neither performance nor training time in any significant way. 
This follows from the observation that for none of the datasets, momentum accounted for more than 1\% of the variance of test set performance.
It should be noted that for TIMIT the interaction between learning rate and momentum accounts for 2.5\% of the total variance, but as with learning rate $\times$ hidden size (cf. \textit{Interaction of Hyperparameters} below) it does not reveal any interpretable structure.
This may be the result of our choice to scale learning rates dependent on momentum (\autoref{sec:arch-training}).
These observations suggest that momentum does not offer substantial benefits when training LSTMs with online stochastic gradient descent. 

\subsubsection*{Analysis of Variance}
\begin{figure}[t]
\centering
\includegraphics[width=0.80\columnwidth]{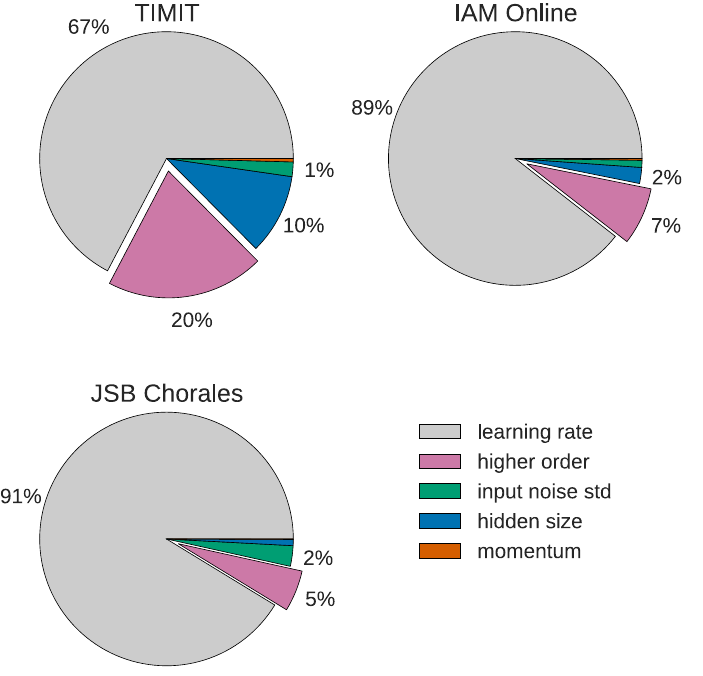}
\caption{Pie charts showing which fraction of variance of the test set performance can be attributed to each of the hyperparameters. The percentage of variance that is due to interactions between multiple parameters is indicated as ``higher order.''} 
\label{fig:variances}

\end{figure}

\autoref{fig:variances} shows what fraction of the test set performance variance can be attributed to different hyperparameters. It is obvious that the learning rate is by far the most important hyperparameter, always accounting for more than two thirds of the variance. The next most important hyperparameter is the hidden layer size, followed by the input noise, leaving the momentum with less than one percent of the variance. Higher order interactions play an important role in the case of TIMIT, but are much less important for the other two data sets. 

\subsubsection*{Interaction of Hyperparameters}
\begin{figure*}
\centering
\subfigure[]{
\includegraphics[width=0.38\textwidth]{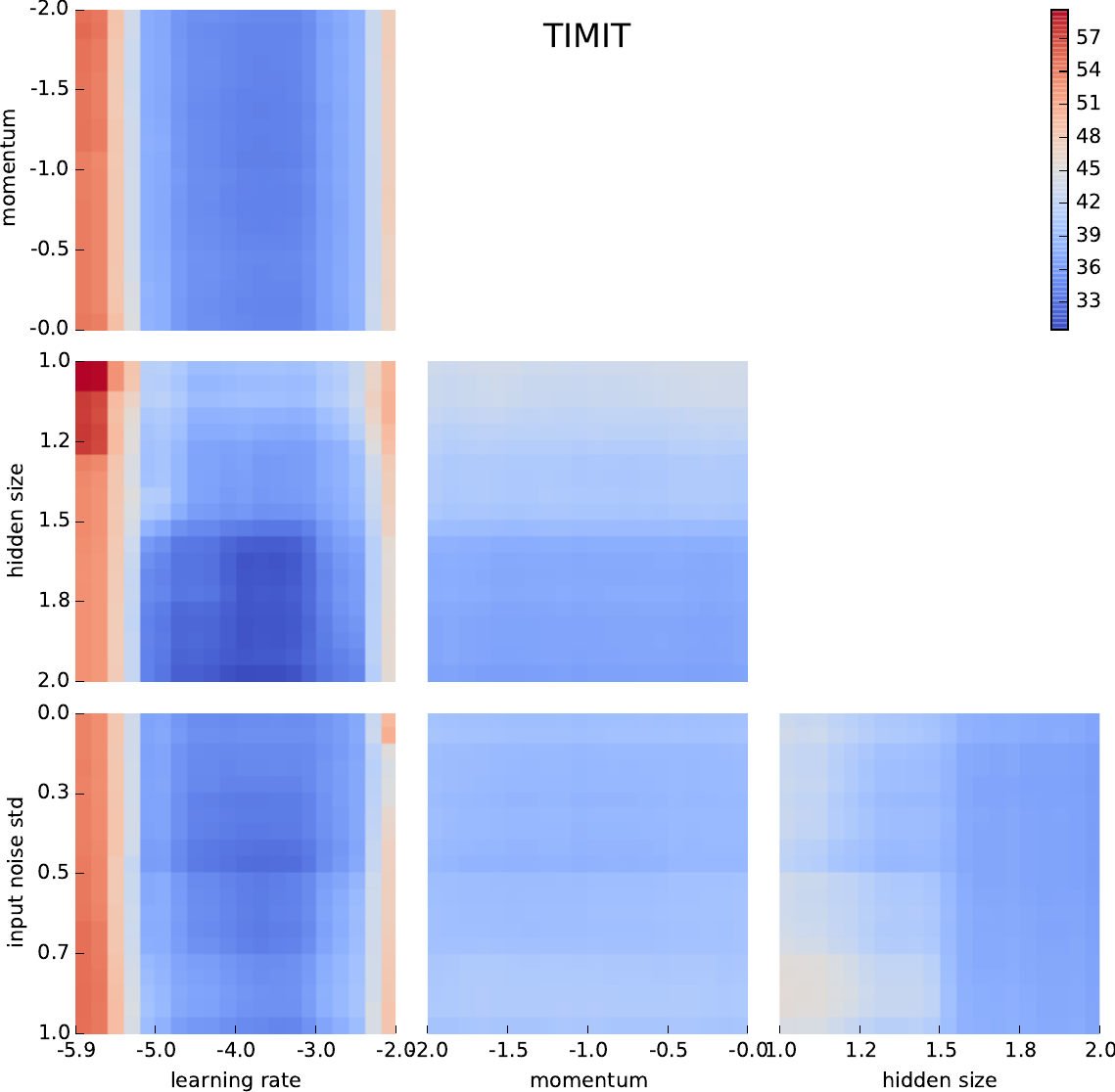}
\includegraphics[width=0.38\textwidth]{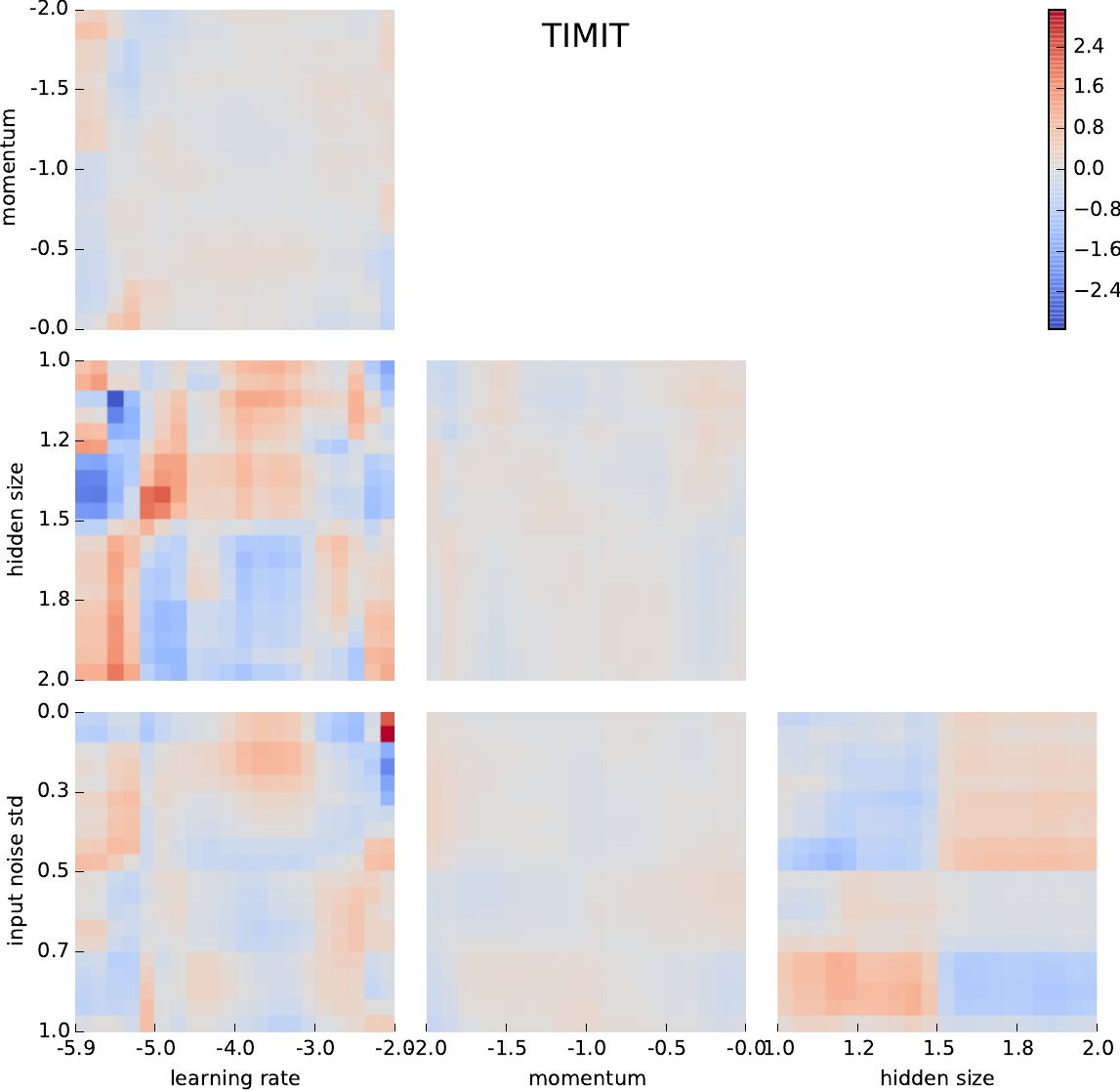}
}
\subfigure[]{
\includegraphics[width=0.38\textwidth]{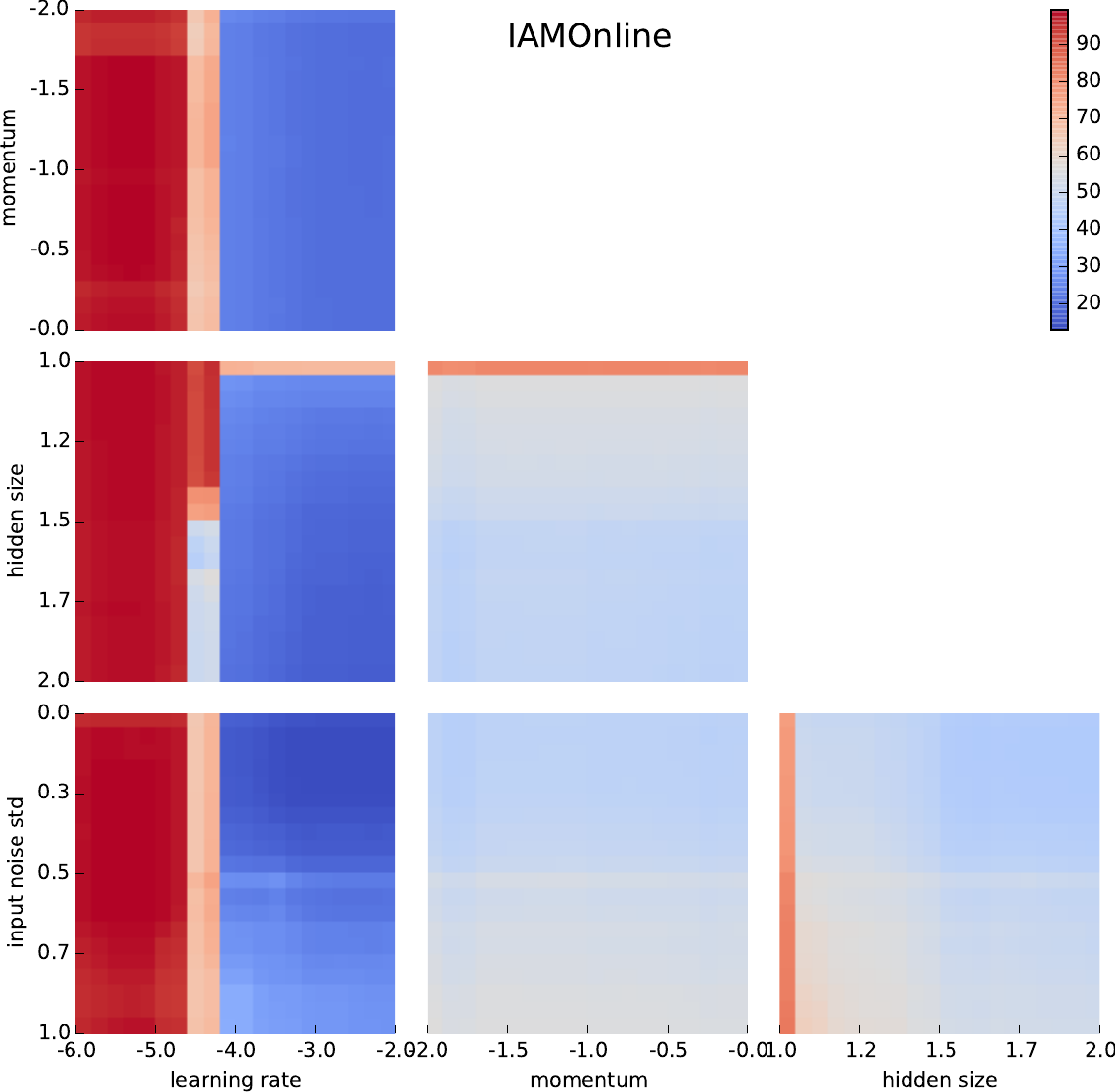}
\includegraphics[width=0.38\textwidth]{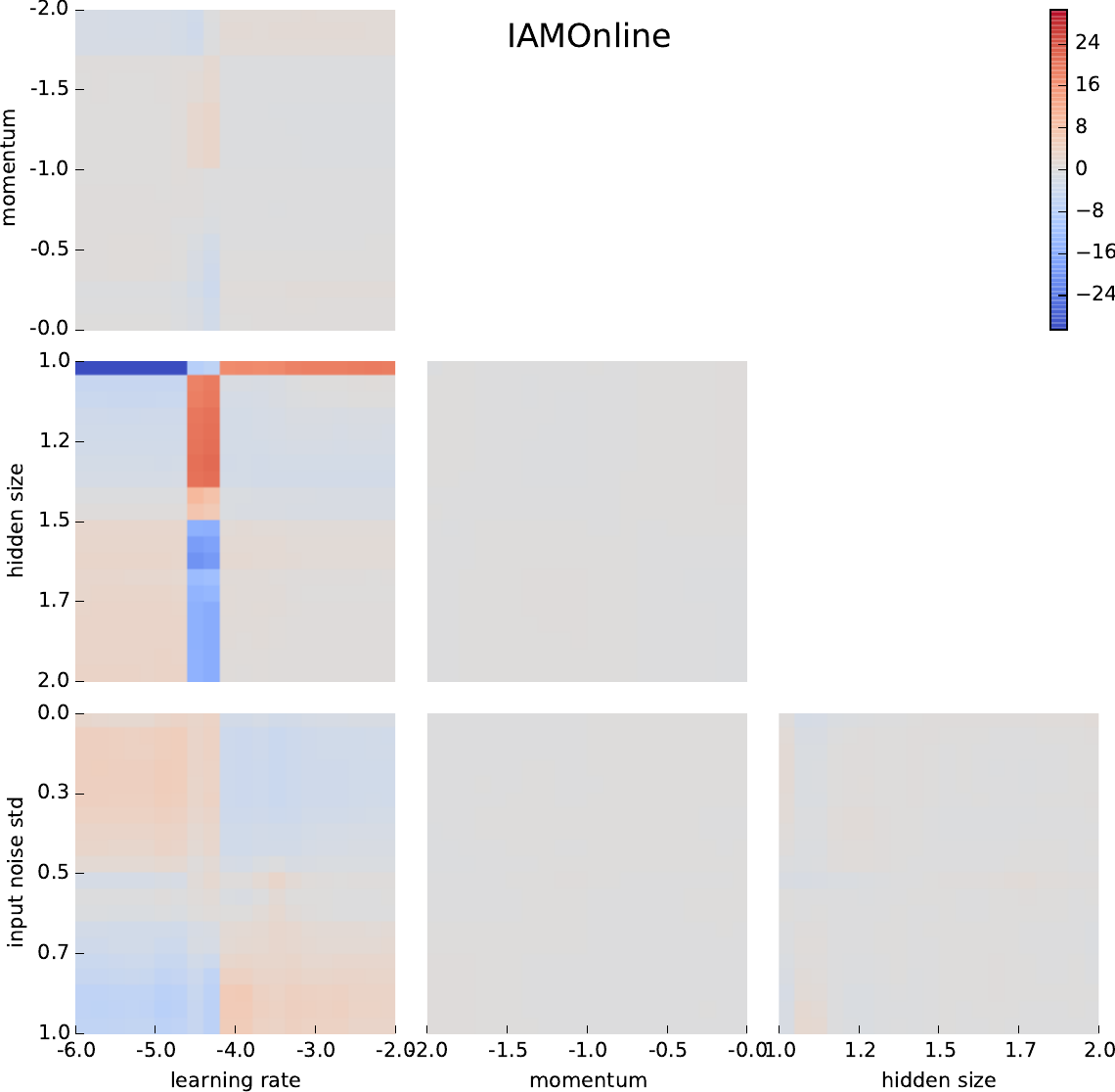}
}
\subfigure[]{
\includegraphics[width=0.38\textwidth]{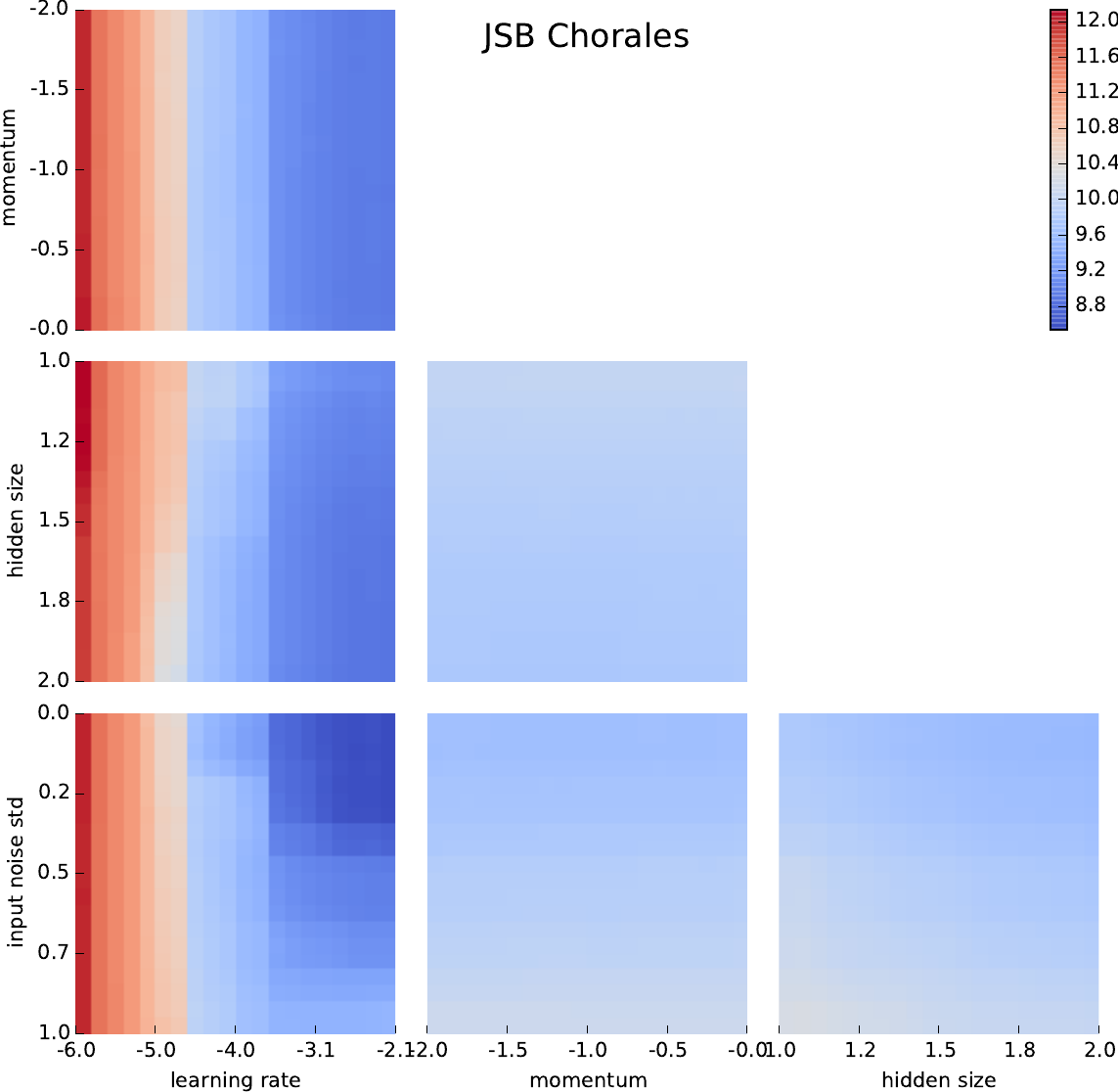}
\includegraphics[width=0.38\textwidth]{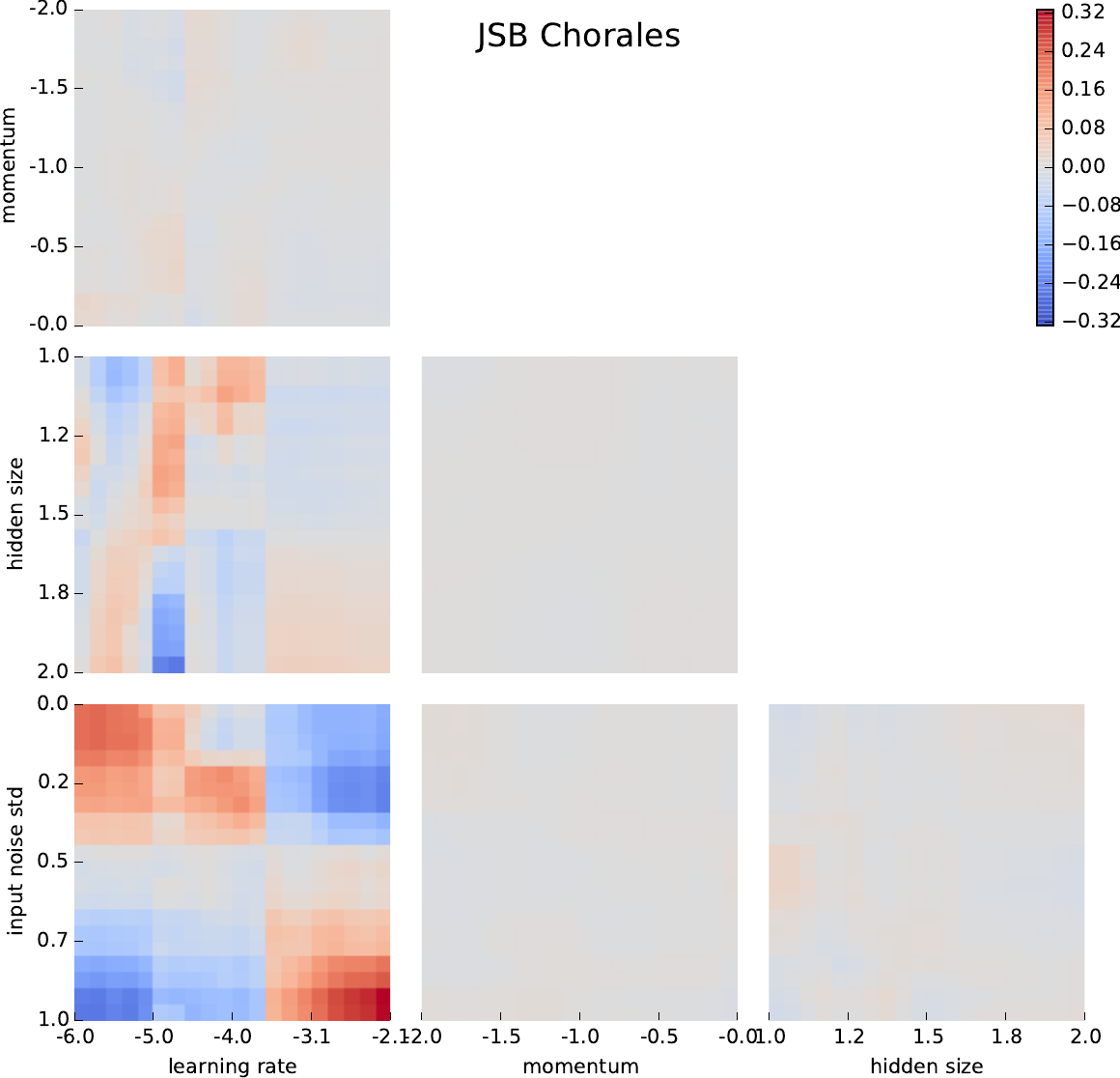}
}
\caption{Total marginal predicted performance for all pairs of hyperparameters (left) and the variation only due to their interaction (right). 
The plot is divided vertically into three subplots, one for every dataset (TIMIT, IAM Online, and JSB Chorales).
The subplots itself are divided horizontally into two parts, each containing a lower triangular matrix of heat maps.
The rows and columns of these matrices represent the different hyperparameters (learning rate, momentum, hidden size, and input noise) and there is one heat map for every combination.
The color encodes the performance as measured by the \emph{Classification Error} for TIMIT,
 \emph{Character Error Rate} for IAM Online and \emph{Negative Log-Likelihood} for the JSB Chorales Dataset.
For all datasets low (blue) is better than high (red).
}
\label{fig:hyper}
\end{figure*}

Some hyperparameters interact with each other resulting in different performance from what could be expected by looking at them individually. 
As shown in \autoref{fig:variances} all these interactions together explain between 5\% and 20\% of the variance in test set performance.
Understanding these interactions might allow us to speed up the search for good combinations of hyperparameters.
To that end we visualize the interaction between all pairs of hyperparameters in \autoref{fig:hyper}.
Each heat map in the left part shows marginal performance for different values of the respective two hyperparameters. 
This is the average performance predicted by the decision forest when marginalizing over all other hyperparameters.
So each one is the 2D version of the performance plots from \autoref{fig:learning_rate} in the paper.

The right side employs the idea of ANOVA to better illustrate the \emph{interaction} between the hyperparameters.  
This means that variance of performance that can be explained by varying a single hyperparameter has been removed. 
In case two hyperparameters do not interact at all (are perfectly independent), that residual would thus be all zero (grey).

For example, looking at the pair \emph{hidden size} and \emph{learning rate} on the left side for the TIMIT dataset, we can see that performance varies strongly along the $x$-axis (learning rate), first decreasing and then increasing again.
This is what we would expect knowing the valley-shape of the learning rate from \autoref{fig:learning_rate}.
Along the $y$-axis (hidden size) performance seems to decrease slightly from top to bottom. 
Again this is roughly what we would expect from the hidden size plot in  \autoref{fig:hidden_size}. 

On the right side of \autoref{fig:hyper} we can see for the same pair of hyperparameters how their interaction differs from the case of them being completely independent. This heat map exhibits less structure, and it may in fact be the case that we would need more samples to properly analyze the interplay between them. However, given our observations so far this might not be worth the effort. In any case, it is clear from the plot on the left that varying the hidden size does not change the region of optimal learning rate.

One clear interaction pattern can be observed in the IAM~Online and JSB datasets between learning rate and input noise.
Here it can be seen that for high learning rates ($\gtrapprox10^{-4}$) lower input noise ($\lessapprox.5$) is better like also observed in the marginals from \autoref{fig:input_noise}.
But this trend reverses for lower learning rates, where higher values of input noise are beneficial. 
Though interesting this is not of any practical relevance because performance is generally bad in that region of low learning rates.
Apart from this, however, it is difficult to discern any regularities in the analyzed hyperparameter interactions. 
We conclude that there is little practical value in attending to the interplay between hyperparameters.
So for practical purposes hyperparameters can be treated as approximately independent and thus optimized separately.


\section{Conclusion}
This paper reports the results of a large scale study on variants of the LSTM architecture. We conclude that the most commonly used LSTM architecture (vanilla LSTM) performs reasonably well on various datasets.
None of the eight investigated modifications significantly improves performance.
However, certain modifications such as coupling the input and forget gates (CIFG) or removing peephole connections (NP) simplified LSTMs in our experiments without significantly decreasing performance.
These two variants are also attractive because they reduce the number of parameters and the computational cost of the LSTM.

The forget gate and the output activation function are the most critical components of the LSTM block.
Removing any of them significantly impairs performance. 
We hypothesize that the output activation function is needed to prevent the unbounded cell state to propagate through the network and destabilize learning. 
This would explain why the LSTM variant GRU can perform reasonably well without it: its cell state is bounded because of the coupling of input and forget gate.

As expected, the learning rate is the most crucial hyperparameter, followed by the network size.
Surprisingly though, the use of momentum was found to be unimportant in our setting of online gradient descent. 
Gaussian noise on the inputs was found to be moderately helpful for TIMIT, but harmful for the other datasets.

The analysis of hyperparameter interactions revealed no apparent structure.
Furthermore, even the highest measured interaction (between learning rate and network size) is quite small.
This implies that for practical purposes the hyperparameters can be treated as approximately independent. 
In particular, the learning rate can be tuned first using a fairly small network, thus saving a lot of experimentation time.

Neural networks can be tricky to use for many practitioners compared to other methods whose properties are already well understood.
This has remained a hurdle for newcomers to the field since a lot of practical choices are based on the intuitions of experts, as well as experiences gained over time. With this study, we have attempted to back some of these intuitions with experimental results.
We have also presented new insights, both on architecture selection and hyperparameter tuning for LSTM networks which have emerged as the method of choice for solving complex sequence learning problems. In future work, we plan to explore more complex modifications of the LSTM architecture.


\bibliography{lstm_study}

\begin{thebibliography}{46}
\providecommand{\natexlab}[1]{#1}
\providecommand{\url}[1]{\texttt{#1}}
\expandafter\ifx\csname urlstyle\endcsname\relax
  \providecommand{\doi}[1]{doi: #1}\else
  \providecommand{\doi}{doi: \begingroup \urlstyle{rm}\Url}\fi

\bibitem[Hochreiter(1991)]{Hochreiter1991}
Sepp Hochreiter.
\newblock \emph{Untersuchungen zu dynamischen neuronalen {Netzen}}.
\newblock Masters {Thesis}, Technische {Universit}\"{a}t {M}\"{u}nchen,
  M\"{u}nchen, 1991.

\bibitem[Hochreiter et~al.(2001)Hochreiter, Bengio, Frasconi, and
  Schmidhuber]{Hochreiter2001}
S.~Hochreiter, Y.~Bengio, P.~Frasconi, and J.~Schmidhuber.
\newblock Gradient flow in recurrent nets: the difficulty of learning long-term
  dependencies.
\newblock In S.~C. Kremer and J.~F. Kolen, editors, \emph{A Field Guide to
  Dynamical Recurrent Neural Networks}. IEEE Press, 2001.

\bibitem[Graves et~al.(2009)Graves, Liwicki, Fernandez, Bertolami, Bunke, and
  Schmidhuber]{Graves2009}
A~Graves, M~Liwicki, S~Fernandez, R~Bertolami, H~Bunke, and J~Schmidhuber.
\newblock A {Novel} {Connectionist} {System} for {Improved} {Unconstrained}
  {Handwriting} {Recognition}.
\newblock \emph{{IEEE} {Transactions} on {Pattern} {Analysis} and {Machine}
  {Intelligence}}, 31\penalty0 (5), 2009.

\bibitem[Pham et~al.(2013)Pham, Bluche, Kermorvant, and Louradour]{Pham2013}
Vu~Pham, Th\'{e}odore Bluche, Christopher Kermorvant, and J\'{e}r\^{o}me
  Louradour.
\newblock Dropout improves {Recurrent} {Neural} {Networks} for {Handwriting}
  {Recognition}.
\newblock \emph{{arXiv}:1312.4569 {[}cs{]}}, November 2013.
\newblock URL \url{http://arxiv.org/abs/1312.4569}.

\bibitem[Doetsch et~al.(2014)Doetsch, Kozielski, and Ney]{Doetsch2014}
Patrick Doetsch, Michal Kozielski, and Hermann Ney.
\newblock Fast and robust training of recurrent neural networks for offline
  handwriting recognition.
\newblock In \emph{14th {International} {Conference} on {Frontiers} in
  {Handwriting} {Recognition}}, 2014.
\newblock URL
  \url{http://people.sabanciuniv.edu/berrin/cs581/Papers/icfhr2014/data/4334a279.pdf}.

\bibitem[Graves(2013)]{Graves2013d}
Alex Graves.
\newblock Generating sequences with recurrent neural networks.
\newblock \emph{{arXiv}:1308.0850 {[}cs{]}}, August 2013.
\newblock URL \url{http://arxiv.org/abs/1308.0850}.

\bibitem[Zaremba et~al.(2014)Zaremba, Sutskever, and Vinyals]{Zaremba2014}
Wojciech Zaremba, Ilya Sutskever, and Oriol Vinyals.
\newblock Recurrent {Neural} {Network} {Regularization}.
\newblock \emph{{arXiv}:1409.2329 {[}cs{]}}, September 2014.
\newblock URL \url{http://arxiv.org/abs/1409.2329}.

\bibitem[Luong et~al.(2014)Luong, Sutskever, Le, Vinyals, and
  Zaremba]{Luong2014}
Thang Luong, Ilya Sutskever, Quoc~V. Le, Oriol Vinyals, and Wojciech Zaremba.
\newblock Addressing the {Rare} {Word} {Problem} in {Neural} {Machine}
  {Translation}.
\newblock \emph{{arXiv} preprint {arXiv}:1410.8206}, 2014.
\newblock URL \url{http://arxiv.org/abs/1410.8206}.

\bibitem[Sak et~al.(2014)Sak, Senior, and Beaufays]{Sak2014}
Hasim Sak, Andrew Senior, and Fran\c{c}oise Beaufays.
\newblock Long short-term memory recurrent neural network architectures for
  large scale acoustic modeling.
\newblock In \emph{Proceedings of the {Annual} {Conference} of {International}
  {Speech} {Communication} {Association} ({INTERSPEECH})}, 2014.
\newblock URL
  \url{http://193.6.4.39/~czap/letoltes/IS14/IS2014/PDF/AUTHOR/IS141304.PDF}.

\bibitem[Fan et~al.(2014)Fan, Qian, Xie, and Soong]{Fan2014}
Yuchen Fan, Yao Qian, Fenglong Xie, and Frank~K. Soong.
\newblock {TTS} synthesis with bidirectional {LSTM} based recurrent neural
  networks.
\newblock In \emph{Proc. {Interspeech}}, 2014.

\bibitem[S{\o}nderby and Winther(2014)]{Sonderby2014}
S{\o}ren~Kaae S{\o}nderby and Ole Winther.
\newblock Protein {Secondary} {Structure} {Prediction} with {Long} {Short}
  {Term} {Memory} {Networks}.
\newblock \emph{{arXiv}:1412.7828 {[}cs, q-bio{]}}, December 2014.
\newblock URL \url{http://arxiv.org/abs/1412.7828}.
\newblock arXiv: 1412.7828.

\bibitem[Marchi et~al.(2014)Marchi, Ferroni, Eyben, Gabrielli, Squartini, and
  Schuller]{Marchi2014}
E.~Marchi, G.~Ferroni, F.~Eyben, L.~Gabrielli, S.~Squartini, and B.~Schuller.
\newblock Multi-resolution linear prediction based features for audio onset
  detection with bidirectional {LSTM} neural networks.
\newblock In \emph{2014 {IEEE} {International} {Conference} on {Acoustics},
  {Speech} and {Signal} {Processing} ({ICASSP})}, pages 2164--2168, May 2014.
\newblock \doi{10.1109/ICASSP.2014.6853982}.

\bibitem[Donahue et~al.(2014)Donahue, Hendricks, Guadarrama, Rohrbach,
  Venugopalan, Saenko, and Darrell]{Donahue2014}
Jeff Donahue, Lisa~Anne Hendricks, Sergio Guadarrama, Marcus Rohrbach,
  Subhashini Venugopalan, Kate Saenko, and Trevor Darrell.
\newblock Long-term {Recurrent} {Convolutional} {Networks} for {Visual}
  {Recognition} and {Description}.
\newblock \emph{{arXiv}:1411.4389 {[}cs{]}}, November 2014.
\newblock URL \url{http://arxiv.org/abs/1411.4389}.
\newblock arXiv: 1411.4389.

\bibitem[Hochreiter and Schmidhuber(1995)]{Hochreiter1995}
Sepp Hochreiter and J\"{u}rgen Schmidhuber.
\newblock Long {Short} {Term} {Memory}.
\newblock Technical {Report} FKI-207-95, Technische {Universit}\"{a}t
  {M}\"{u}nchen, M\"{u}nchen, August 1995.
\newblock URL
  \url{http://citeseerx.ist.psu.edu/viewdoc/summary?doi=10.1.1.51.3117}.

\bibitem[Hochreiter and Schmidhuber(1997)]{Hochreiter1997}
Sepp Hochreiter and J\"{u}rgen Schmidhuber.
\newblock Long {Short}-{Term} {Memory}.
\newblock \emph{Neural {Computation}}, 9\penalty0 (8):\penalty0 1735--1780,
  November 1997.
\newblock ISSN 0899-7667.
\newblock \doi{10.1162/neco.1997.9.8.1735}.
\newblock URL \url{http://www.bioinf.jku.at/publications/older/2604.pdf}.

\bibitem[Anderson(1953)]{Anderson1953}
R.~L. Anderson.
\newblock Recent {Advances} in {Finding} {Best} {Operating} {Conditions}.
\newblock \emph{Journal of the American Statistical Association}, 48\penalty0
  (264):\penalty0 789--798, December 1953.
\newblock ISSN 0162-1459.
\newblock \doi{10.2307/2281072}.
\newblock URL \url{http://www.jstor.org/stable/2281072}.

\bibitem[Solis and Wets(1981)]{Solis1981}
Francisco~J. Solis and Roger J.-B. Wets.
\newblock Minimization by {Random} {Search} {Techniques}.
\newblock \emph{Mathematics of Operations Research}, 6\penalty0 (1):\penalty0
  19--30, February 1981.
\newblock ISSN 0364-765X.
\newblock \doi{10.1287/moor.6.1.19}.
\newblock URL \url{http://pubsonline.informs.org/doi/abs/10.1287/moor.6.1.19}.

\bibitem[Bergstra and Bengio(2012)]{Bergstra2012}
James Bergstra and Yoshua Bengio.
\newblock Random search for hyper-parameter optimization.
\newblock \emph{The {Journal} of {Machine} {Learning} {Research}}, 13\penalty0
  (1):\penalty0 281--305, 2012.
\newblock URL \url{http://dl.acm.org/citation.cfm?id=2188395}.

\bibitem[Hutter et~al.(2014)Hutter, Hoos, and Leyton-Brown]{Hutter2014}
Frank Hutter, Holger Hoos, and Kevin Leyton-Brown.
\newblock An {Efficient} {Approach} for {Assessing} {Hyperparameter}
  {Importance}.
\newblock pages 754--762, 2014.
\newblock URL \url{http://jmlr.org/proceedings/papers/v32/hutter14.html}.

\bibitem[Graves and Schmidhuber(2005)]{Graves2005}
Alex Graves and J\"{u}rgen Schmidhuber.
\newblock Framewise phoneme classification with bidirectional {LSTM} and other
  neural network architectures.
\newblock \emph{Neural {Networks}}, 18\penalty0 (5{\textendash}6):\penalty0
  602--610, July 2005.
\newblock ISSN 0893-6080.
\newblock \doi{10.1016/j.neunet.2005.06.042}.
\newblock URL
  \url{http://www.sciencedirect.com/science/article/pii/S0893608005001206}.

\bibitem[Gers et~al.(1999)Gers, Schmidhuber, and Cummins]{Gers1999}
Felix~A. Gers, J\"{u}rgen Schmidhuber, and Fred Cummins.
\newblock Learning to forget: {Continual} prediction with {LSTM}.
\newblock In \emph{Artificial {Neural} {Networks}, 1999. {ICANN} 99. {Ninth}
  {International} {Conference} on ({Conf}. {Publ}. {No}. 470)}, volume~2, pages
  850--855, 1999.

\bibitem[Gers and Schmidhuber(2000)]{Gers2000}
Felix~A. Gers and J\"{u}rgen Schmidhuber.
\newblock Recurrent nets that time and count.
\newblock In \emph{Neural {Networks}, 2000. {IJCNN} 2000, {Proceedings} of the
  {IEEE}-{INNS}-{ENNS} {International} {Joint} {Conference} on}, volume~3,
  pages 189--194. {IEEE}, 2000.
\newblock ISBN 0769506194.

\bibitem[Robinson and Fallside(1987)]{Robinson1987}
AJ~Robinson and Frank Fallside.
\newblock \emph{The utility driven dynamic error propagation network}.
\newblock University of Cambridge Department of Engineering, 1987.

\bibitem[Williams(1989)]{Williams1989}
R.~J. Williams.
\newblock Complexity of exact gradient computation algorithms for recurrent
  neural networks.
\newblock Technical Report Technical Report NU-CCS-89-27, Boston: Northeastern
  University, College of Computer Science, 1989.

\bibitem[Werbos(1988)]{Werbos1988}
P.~J. Werbos.
\newblock Generalization of backpropagation with application to a recurrent gas
  market model.
\newblock \emph{Neural Networks}, 1, 1988.

\bibitem[Garofolo et~al.(1993)Garofolo, Lamel, Fisher, Fiscus, Pallett, and
  Dahlgren]{Garofolo1993}
JS~Garofolo, LF~Lamel, WM~Fisher, JG~Fiscus, DS~Pallett, and NL~Dahlgren.
\newblock {DARPA} {TIMIT} {Acoustic}-{Phonetic} {Continuous} {Speech} {Corpus}
  {CD}-{ROM}.
\newblock \emph{National {Institute} of {Standards} and {Technology}}, {NTIS}
  {Order} {No} {PB}91-505065, 1993.

\bibitem[Gers et~al.(2002)Gers, P\'{e}rez-Ortiz, Eck, and
  Schmidhuber]{Gers2002}
Felix~A. Gers, Juan~Antonio P\'{e}rez-Ortiz, Douglas Eck, and J\"{u}rgen
  Schmidhuber.
\newblock {DEFK}-{LSTM}.
\newblock In \emph{{ESANN} 2002, Proceedings of the 10th Eurorean Symposium on
  Artificial Neural Networks}, 2002.

\bibitem[Schmidhuber et~al.(2007)Schmidhuber, Wierstra, Gagliolo, and
  Gomez]{Schmidhuber2007}
J~Schmidhuber, D~Wierstra, M~Gagliolo, and F~J Gomez.
\newblock Training {Recurrent} {Networks} by {EVOLINO}.
\newblock \emph{Neural {Computation}}, 19\penalty0 (3):\penalty0 757--779,
  2007.

\bibitem[Bayer et~al.(2009)Bayer, Wierstra, Togelius, and
  Schmidhuber]{Bayer2009}
Justin Bayer, Daan Wierstra, Julian Togelius, and J\"{u}rgen Schmidhuber.
\newblock Evolving memory cell structures for sequence learning.
\newblock In \emph{Artificial {Neural} {Networks}{\textendash}{ICANN} 2009},
  pages 755--764. Springer, 2009.
\newblock URL
  \url{http://link.springer.com/chapter/10.1007/978-3-642-04277-5_76}.

\bibitem[Jozefowicz et~al.(2015)Jozefowicz, Zaremba, and
  Sutskever]{Jozefowicz2015}
Rafal Jozefowicz, Wojciech Zaremba, and Ilya Sutskever.
\newblock An empirical exploration of recurrent network architectures.
\newblock In \emph{Proceedings of the 32nd International Conference on Machine
  Learning (ICML-15)}, pages 2342--2350, 2015.

\bibitem[Otte et~al.(2014)Otte, Liwicki, and Zell]{Otte2014}
Sebastian Otte, Marcus Liwicki, and Andreas Zell.
\newblock Dynamic {Cortex} {Memory}: {Enhancing} {Recurrent} {Neural}
  {Networks} for {Gradient}-{Based} {Sequence} {Learning}.
\newblock In \emph{Artificial {Neural} {Networks} and {Machine} {Learning}
  {\textendash} {ICANN} 2014}, number 8681 in Lecture {Notes} in {Computer}
  {Science}, pages 1--8. Springer {International} {Publishing}, September 2014.
\newblock ISBN 978-3-319-11178-0, 978-3-319-11179-7.
\newblock URL
  \url{http://link.springer.com/chapter/10.1007/978-3-319-11179-7_1}.

\bibitem[Cho et~al.(2014)Cho, van Merrienboer, Gulcehre, Bougares, Schwenk, and
  Bengio]{Cho2014}
Kyunghyun Cho, Bart van Merrienboer, Caglar Gulcehre, Fethi Bougares, Holger
  Schwenk, and Yoshua Bengio.
\newblock Learning {Phrase} {Representations} using {RNN} {Encoder}-{Decoder}
  for {Statistical} {Machine} {Translation}.
\newblock \emph{{arXiv} preprint {arXiv}:1406.1078}, 2014.
\newblock URL \url{http://arxiv.org/abs/1406.1078}.

\bibitem[Chung et~al.(2014)Chung, Gulcehre, Cho, and Bengio]{Chung2014}
Junyoung Chung, Caglar Gulcehre, KyungHyun Cho, and Yoshua Bengio.
\newblock Empirical {Evaluation} of {Gated} {Recurrent} {Neural} {Networks} on
  {Sequence} {Modeling}.
\newblock \emph{{arXiv}:1412.3555 {[}cs{]}}, December 2014.
\newblock URL \url{http://arxiv.org/abs/1412.3555}.

\bibitem[Crystal(2011)]{Crystal2011}
David Crystal.
\newblock \emph{Dictionary of linguistics and phonetics}, volume~30.
\newblock John Wiley \& Sons, 2011.

\bibitem[Mermelstein(1976)]{Mermelstein1976}
P.~Mermelstein.
\newblock Distance measures for speech recognition: Psychological and
  instrumental.
\newblock In C.~H. Chen, editor, \emph{Pattern Recognition and Artificial
  Intelligence}, pages 374--388. Academic Press, New York, 1976.

\bibitem[Graves(2008)]{Graves2008a}
Alexander Graves.
\newblock \emph{Supervised {{Sequence Labelling}} with {{Recurrent Neural
  Networks}}}.
\newblock Ph.d., The Technical University of Munich, 2008.

\bibitem[Halberstadt(1998)]{Halberstadt1998}
Andrew~K. Halberstadt.
\newblock \emph{Heterogeneous acoustic measurements and multiple classifiers
  for speech recognition}.
\newblock PhD thesis, Massachusetts {Institute} of {Technology}, 1998.

\bibitem[Liwicki and Bunke(2005)]{Liwicki2005}
Marcus Liwicki and Horst Bunke.
\newblock {IAM}-{OnDB}-an on-line {English} sentence database acquired from
  handwritten text on a whiteboard.
\newblock In \emph{Document {Analysis} and {Recognition}, 2005. {Proceedings}.
  {Eighth} {International} {Conference} on}, pages 956--961. {IEEE}, 2005.

\bibitem[Graves et~al.(2006)Graves, Fern\'{a}ndez, Gomez, and
  Schmidhuber]{Graves2006}
Alex Graves, Santiago Fern\'{a}ndez, Faustino Gomez, and J\"{u}rgen
  Schmidhuber.
\newblock Connectionist temporal classification: labelling unsegmented sequence
  data with recurrent neural networks.
\newblock In \emph{Proceedings of the 23rd international conference on
  {Machine} learning}, pages 369--376, 2006.
\newblock URL \url{http://dl.acm.org/citation.cfm?id=1143891}.

\bibitem[Allan and Williams(2005)]{Allan2005}
Moray Allan and Christopher~KI Williams.
\newblock Harmonising chorales by probabilistic inference.
\newblock \emph{Advances in neural information processing systems},
  17:\penalty0 25--32, 2005.

\bibitem[Boulanger-Lewandowski et~al.(2012)Boulanger-Lewandowski, Bengio, and
  Vincent]{Boulanger-Lewandowski2012}
Nicolas Boulanger-Lewandowski, Yoshua Bengio, and Pascal Vincent.
\newblock Modeling {Temporal} {Dependencies} in {High}-{Dimensional}
  {Sequences}: {Application} to {Polyphonic} {Music} {Generation} and
  {Transcription}.
\newblock pages 1159--1166, 2012.
\newblock URL \url{http://icml.cc/discuss/2012/590.html}.

\bibitem[Sutskever et~al.(2013)Sutskever, Martens, Dahl, and
  Hinton]{Sutskever2013}
Ilya Sutskever, James Martens, George Dahl, and Geoffrey Hinton.
\newblock On the importance of initialization and momentum in deep learning.
\newblock In \emph{{JMLR}}, pages 1139--1147, 2013.
\newblock URL \url{http://jmlr.org/proceedings/papers/v28/sutskever13.html}.

\bibitem[Snoek et~al.(2012)Snoek, Larochelle, and Adams]{Snoek2012}
Jasper Snoek, Hugo Larochelle, and Ryan~P Adams.
\newblock Practical {Bayesian} {Optimization} of {Machine} {Learning}
  {Algorithms}.
\newblock In F.~Pereira, C.~J.~C. Burges, L.~Bottou, and K.~Q. Weinberger,
  editors, \emph{Advances in {Neural} {Information} {Processing} {Systems} 25},
  pages 2951--2959. Curran {Associates}, {Inc}., 2012.

\bibitem[Hutter et~al.(2011)Hutter, Hoos, and Leyton-Brown]{Hutter2011}
F.~Hutter, H.~H. Hoos, and K.~Leyton-Brown.
\newblock Sequential {Model}-{Based} {Optimization} for {General} {Algorithm}
  {Configuration}.
\newblock In \emph{Proc. of {LION}-5}, pages 507--523, 2011.

\bibitem[Graves et~al.(2008)Graves, Liwicki, Bunke, Schmidhuber, and
  Fern\'{a}ndez]{Graves2008}
Alex Graves, Marcus Liwicki, Horst Bunke, J\"{u}rgen Schmidhuber, and Santiago
  Fern\'{a}ndez.
\newblock Unconstrained on-line handwriting recognition with recurrent neural
  networks.
\newblock In \emph{Advances in {Neural} {Information} {Processing} {Systems}},
  pages 577--584, 2008.

\bibitem[Hooker(2007)]{Hooker2007}
Giles Hooker.
\newblock Generalized {Functional} {ANOVA} {Diagnostics} for
  {High}-{Dimensional} {Functions} of {Dependent} {Variables}.
\newblock \emph{Journal of {Computational} and {Graphical} {Statistics}},
  16\penalty0 (3):\penalty0 709--732, September 2007.
\newblock ISSN 1061-8600, 1537-2715.
\newblock \doi{10.1198/106186007X237892}.
\newblock URL
  \url{http://www.tandfonline.com/doi/abs/10.1198/106186007X237892}.

\end{thebibliography}
\bibliographystyle{unsrtnat}


\begin{IEEEbiography}[{\includegraphics[width=1in,height=1.25in,clip,keepaspectratio]{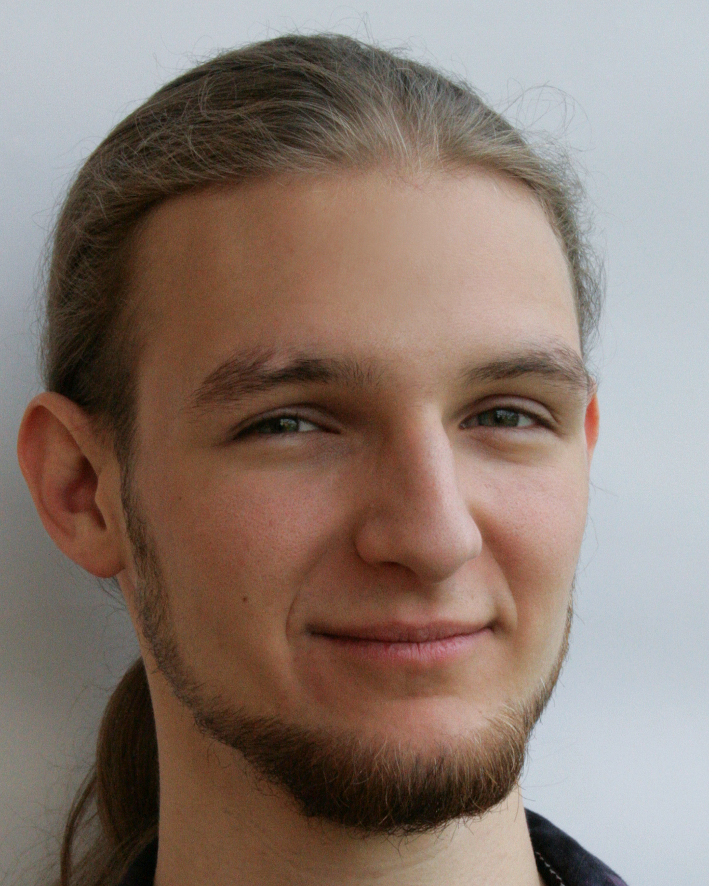}}]{Klaus Greff}
received his Diploma in Computer Science from the University
of Kaiserslautern, Germany in 2011. Currently he is pursuing his PhD at
IDSIA in Lugano, Switzerland, under the supervision of Prof. J\"urgen
Schmidhuber in the field of Machine Learning. His research interests
include Sequence Learning and Recurrent Neural Networks.
\end{IEEEbiography}

\begin{IEEEbiography}[{\includegraphics[width=1in,height=1.25in,clip,keepaspectratio]{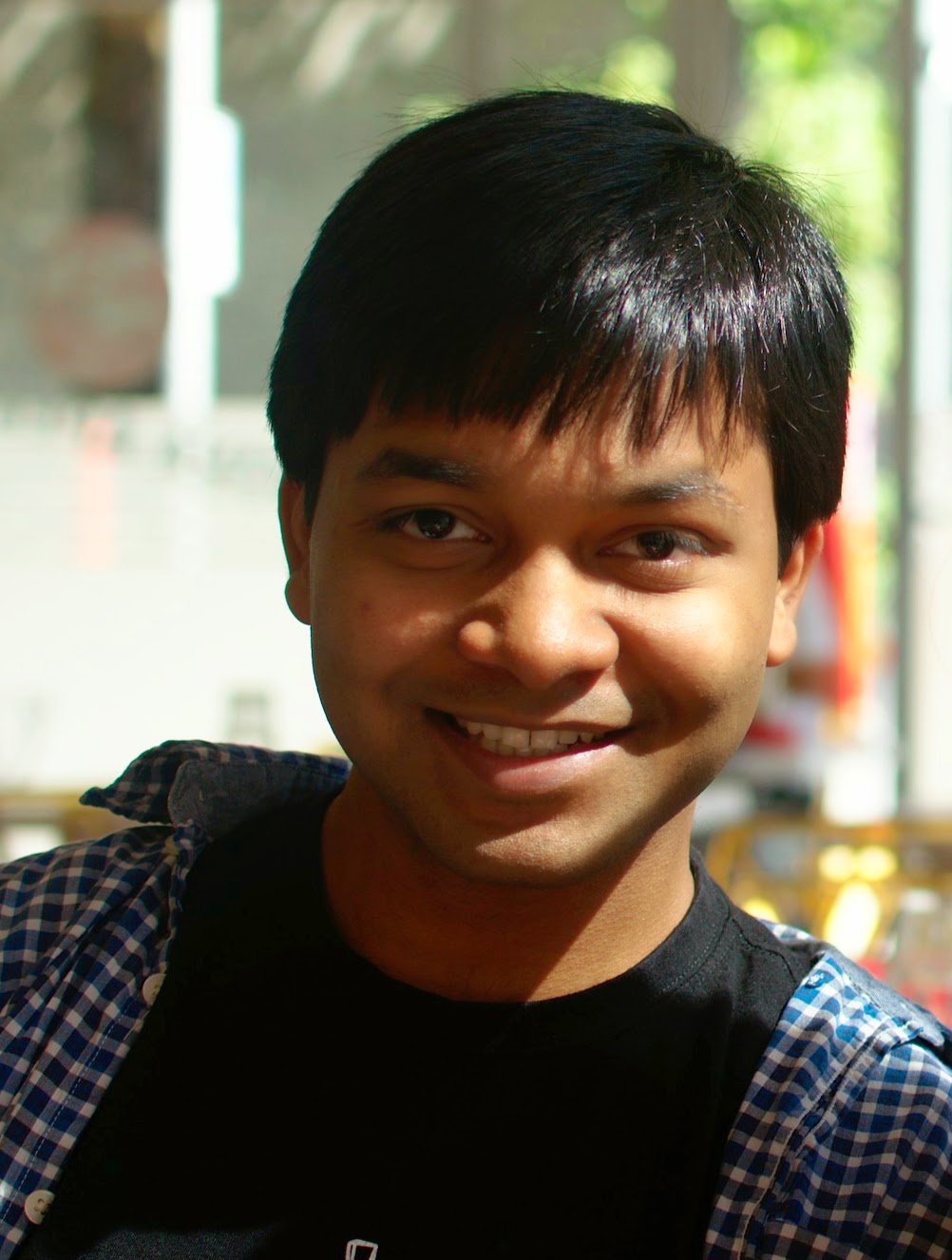}}]{Rupesh Srivastava}
is a PhD student at IDSIA \& USI in Switzerland, supervised by Prof. Jürgen Schmidhuber. 
He currently works on understanding and improving neural network architectures. 
In particular, he has worked on understanding the beneficial properties of local competition in neural networks, and new architectures which allow gradient-based training of extremely deep networks. 
In the past, Rupesh worked on reliability based design optimization using evolutionary algorithms at the Kanpur Genetic Algorithms Laboratory, supervised by Prof. Kalyanmoy Deb for his Masters thesis.
\end{IEEEbiography}

\begin{IEEEbiography}[{\includegraphics[width=1in,height=1.25in,clip,keepaspectratio]{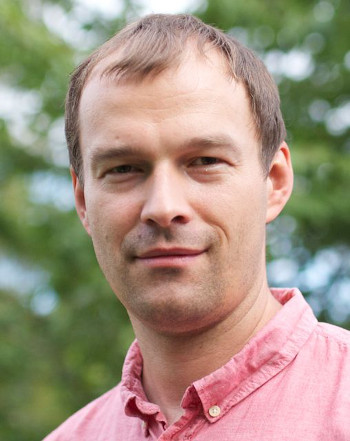}}]{Jan Koutn\'ik}
received his Ph.D. in computer science from the Czech Technical University in Prague in 2008. He works as machine learning researcher at The Swiss AI Lab IDSIA. His research is mainly focused on artificial neural networks, recurrent neural networks, evolutionary algorithms and deep-learning applied to reinforcement learning, control problems, image classification, handwriting and speech recognition. 
\end{IEEEbiography}

\begin{IEEEbiography}[{\includegraphics[width=1in,height=1in,clip,keepaspectratio]{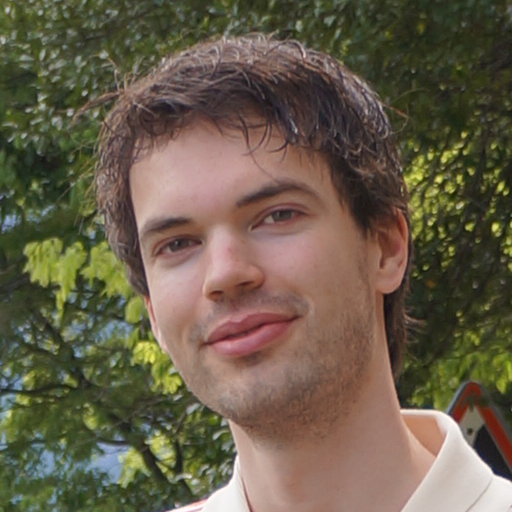}}]{Bas R. Steunebrink}
is a postdoctoral researcher at the Swiss AI lab IDSIA. He received his PhD in 2010 from Utrecht University, the Netherlands. Bas's interests and expertise include Artificial General Intelligence (AGI), cognitive robotics, machine learning, resource-constrained control, and affective computing.
\end{IEEEbiography}

\begin{IEEEbiography}[{\includegraphics[width=1in,height=1.25in,clip,keepaspectratio]{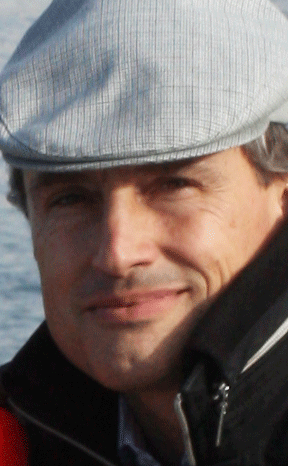}}]{J\"urgen Schmidhuber}
is Professor of Artificial Intelligence (AI) at USI in Switzerland. 
He has pioneered self-improving general problem solvers since 1987, and Deep Learning Neural Networks (NNs) since 1991. 
The recurrent NNs (RNNs) developed by his research groups at the Swiss AI Lab IDSIA \& USI \& SUPSI \& TU Munich were the first RNNs to win official international contests. 
They have helped to revolutionize connected handwriting recognition, speech recognition, machine translation, optical character recognition, image caption generation, and are now in use at Google, Apple, Microsoft, IBM, Baidu, and many other companies. 
IDSIA's Deep Learners were also the first to win object detection and image segmentation contests, and achieved the world's first superhuman visual classification results, winning nine international competitions in machine learning \& pattern recognition (more than any other team). 
They also were the first to learn control policies directly from high-dimensional sensory input using reinforcement learning. 
His research group also established the field of mathematically rigorous universal AI and optimal universal problem solvers. 
His formal theory of creativity \& curiosity \& fun explains art, science, music, and humor. 
He also generalized algorithmic information theory and the many-worlds theory of physics, and introduced the concept of Low-Complexity Art, the information age's extreme form of minimal art. 
Since 2009 he has been member of the European Academy of Sciences and Arts. 
He has published 333 peer-reviewed papers, earned seven best paper/best video awards, and is recipient of the 2013 Helmholtz Award of the International Neural Networks Society and the 2016 IEEE Neural Networks Pioneer Award. 
\end{IEEEbiography}

\end{document}